\documentclass[sigconf]{acmart}

\usepackage{subfig}
\usepackage{bm}
\usepackage{hyperref}
\usepackage{xcolor}
\usepackage{paralist}

\AtBeginDocument{%
  }

\copyrightyear{2024}
\acmYear{2024}
\setcopyright{acmlicensed}
\acmConference[MM '24]{Proceedings of the 32nd ACM International Conference on Multimedia}{October 28-November 1, 2024}{Melbourne, VIC, Australia}
\acmBooktitle{Proceedings of the 32nd ACM International Conference on Multimedia (MM '24), October 28-November 1, 2024, Melbourne, VIC, Australia}
\acmDOI{10.1145/3664647.3681404}
\acmISBN{979-8-4007-0686-8/24/10}

\begin{document}

\title{Dynamic Evidence Decoupling for Trusted Multi-view Learning}


\author{Ying Liu}
\orcid{0000-0002-3117-9926}
\affiliation{%
  \department{School of Computer Science and Technology}
  \institution{Xidian University}
  \city{Xi'an}
  \country{China}}
\email{ying210281@163.com}

\author{Lihong Liu}
\orcid{0009-0003-4938-2126}
\affiliation{%
  \department{School of Computer Science and Technology}
  \institution{Xidian University}
  \city{Xi'an}
  \country{China}}
\email{21lsonelh818@stu.xidian.edu.cn}

\author{Cai Xu}
\orcid{0000-0002-7191-7348}
\affiliation{%
  \department{School of Computer Science and Technology}
  \institution{Xidian University}
  \city{Xi'an}
  \country{China}}
\email{cxu@xidian.edu.cn}
\authornote{Corresponding Author.}

\author{Xiangyu Song}
\orcid{0000-0002-5550-6354}
\affiliation{%
  \department{New Network Research Division}
  \institution{Pengcheng Laboratory}
  \city{Shenzhen}
  \country{China}}
\email{songxy02@pcl.ac.cn}

\author{Ziyu Guan}
\orcid{0000-0003-2413-4698}
\affiliation{%
  \department{School of Computer Science and Technology}
  \institution{Xidian University}
  \city{Xi'an}
  \country{China}}
\email{ziyuguan@xidian.edu.cn}

\author{Wei Zhao}
\orcid{0000-0002-9767-1323}
\affiliation{%
  \department{School of Computer Science and Technology}
  \institution{Xidian University}
  \city{Xi'an}
  \country{China}}
\email{ywzhao@mail.xidian.edu.cn}


\renewcommand{\shortauthors}{Ying Liu et al.}

\begin{abstract}
  Multi-view learning methods often focus on improving decision accuracy, while neglecting the decision uncertainty, limiting their suitability for safety-critical applications. To mitigate this, researchers propose trusted multi-view learning methods that estimate classification probabilities and uncertainty by learning the class distributions for each instance. However, these methods assume that the data from each view can effectively differentiate all categories, ignoring the semantic vagueness phenomenon in real-world multi-view data. Our findings demonstrate that this phenomenon significantly suppresses the learning of view-specific evidence in existing methods. We propose a Consistent and Complementary-aware trusted Multi-view Learning (CCML) method to solve this problem. We first construct view opinions using evidential deep neural networks, which consist of belief mass vectors and uncertainty estimates. Next, we dynamically decouple the consistent and complementary evidence. The consistent evidence is derived from the shared portions across all views, while the complementary evidence is obtained by averaging the differing portions across all views. We ensure that the opinion constructed from the consistent evidence strictly aligns with the ground-truth category. For the opinion constructed from the complementary evidence, we allow it for potential vagueness in the evidence. We compare CCML with state-of-the-art baselines on one synthetic and six real-world datasets. The results validate the effectiveness of the dynamic evidence decoupling strategy and show that CCML significantly outperforms baselines on accuracy and reliability. The code is released at https://github.com/Lihong-Liu/CCML.
\end{abstract}





\maketitle

\section{Introduction}
In real-world scenarios, different data modalities or features could be treated as multiple views. For example, in autonomous vehicle systems, cameras and lidars collect images and points; in the field of healthcare, a patient's comprehensive condition can be assessed through multiple types of examinations. Multi-view learning aims to synthesize both consistent and complementary information from these multiple views, leading to a more comprehensive understanding of the data \cite{wei2024video, liu2022disentangled, dong2023cross}. It has generated significant and wide-ranging influence across multiple research areas, including classification \cite{liang2021af}, semi-supervised learning \cite{10440511}, clustering \cite{huang2020auto, sun2021scalable}, retrieval \cite{zhen2019deep} and large language models \cite{min2023recent}.

\begin{figure}
    \centering
    \includegraphics[width=0.47\textwidth]{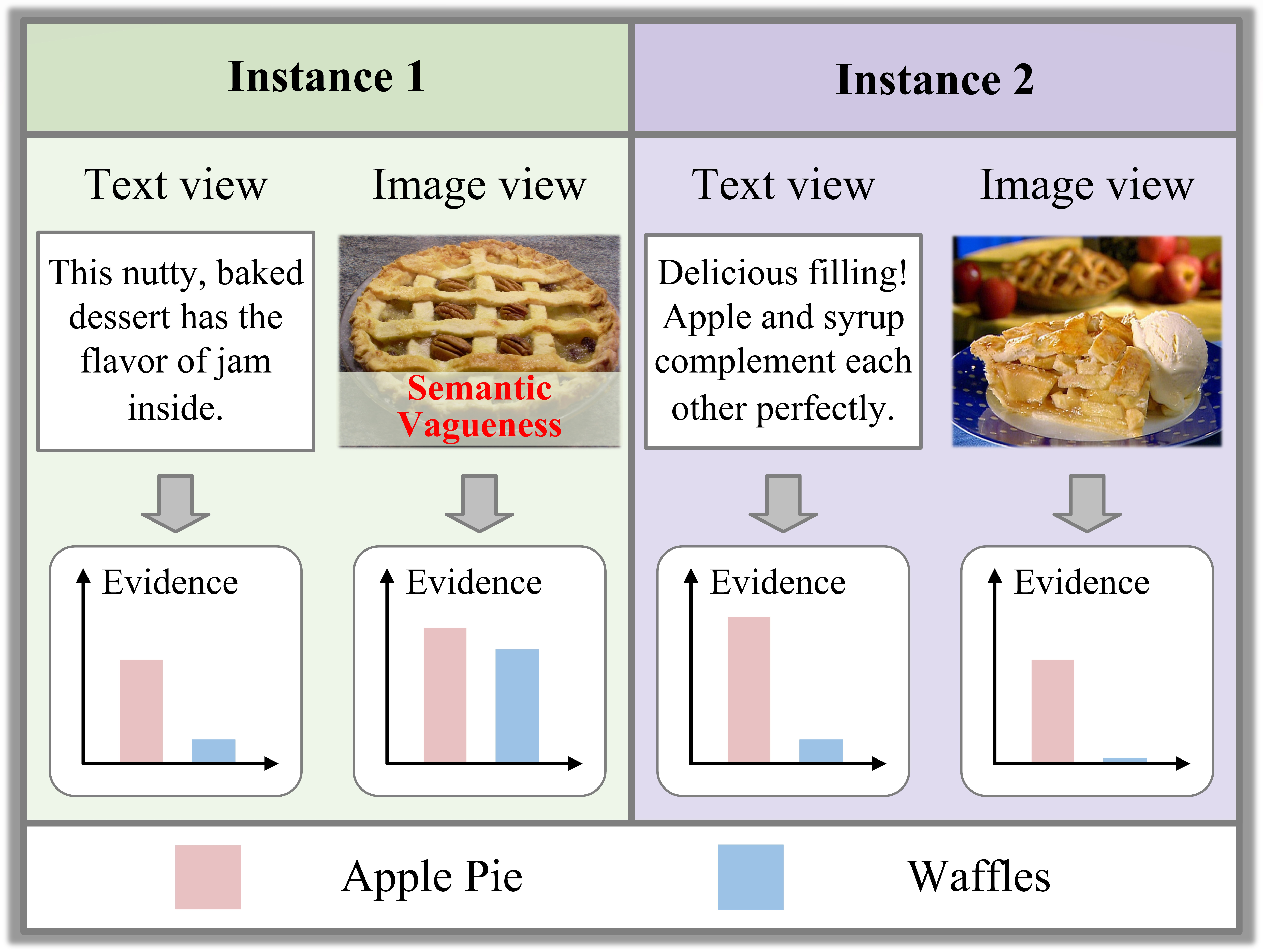}
    \caption{Visualization of the dynamic semantic vagueness phenomenon. The ground-truth category of the first instance is ``apple pie''. When considering the image view alone, it becomes challenging to differentiate between the categories ``waffles'' and ``apple pie''. For the second instance, both views provide explicit differentiation between the categories. }
    \label{fig:introduction}
\end{figure}

Most existing multi-view learning methods primarily emphasize enhancing decision accuracy, often overlooking the crucial aspect of decision uncertainty. This limitation significantly restricts the applicability of multi-view learning in safety-critical scenarios, such as autonomous vehicle systems. Therefore, to predict the reliability of the results and further boost performance, researchers have proposed many multi-view uncertainty quantification methods in recent years. The pioneering work \cite{han2021tmc}, Trusted Multi-view Classification (TMC), calculates and aggregates the evidence \cite{sensoy2018evidential} of all views. TMC utilizes this aggregated evidence to parameterize the class distributions, enabling the estimation of class probabilities and uncertainties. In order to train the model effectively, TMC requires the estimated class probabilities to align with the ground-truth labels. Building upon this research, researchers have proposed novel evidence aggregation paradigms such as sum \cite{liu2022trusted}, (weighted) average \cite{xu2024reliable}, element-wise dot product \cite{han2022trusted}, etc. These methods enhance the reliability in the presence of various challenges, such as feature noise \cite{gan2021brain, qin2022deep, zhou2023calm}, incomplete views \cite{xie2023exploring}, etc.

%Many traditional multi-view deep learning methods have accomplished multi-view classification tasks and can obtain the accuracy of classification. However, due to the differences or variations in the quality of different views in real-world scenarios, we need to assess the reliability of the obtained results in many cases, such as in high-risk medical diagnoses. The limitations of these traditional methods make it difficult for them to solve this problem. 

%Therefore, to predict the reliability of the results, many multi-view uncertainty-aware methods have been proposed, with Trusted Multi-view Classification (TMC) \cite{han2021tmc} based on Evidential Deep Learning (EDL) \cite{sensoy2018evidential} being representative. TMC converts the evidence obtained from the EDL method into opinions through the subjective logic framework and uses the DS evidence theory to merge opinions from different views. Trusted Multi-View Deep Learning with Opinion Aggregation (TMDL-OA) \cite{liu2022trusted}, by maximization of consistency across views to realize a trusted aggregation strategy and reduce the overall uncertainty after aggregation. Safe Multi-View Deep Classification (SMDC) \cite{liu2023safe} can guarantee that the classification performance does not deteriorate when fusing multiple views.

Regrettably, the evidence aggregation paradigms in these methods rely on an assumption: the data of each view can distinguish all categories. However, real-world multi-view data exhibit the \textit{semantic vagueness phenomenon}, which means that one view may exhibit ambiguity or uncertainty in its categorization. For example, as shown in Figure \ref{fig:introduction}, the ground-truth category of the first instance is ``apple pie'', while the image view is difficult to differentiate between the categories ``waffles'' and ``apple pie''. For this view, existing evidence aggregation paradigms encourage only the evidence of the category ``apple pie'' is large due to the common evidence complying with this. This overlooks the fact that the evidence for the category ``waffles'' is also substantial in the data, which would significantly impact the overall learning process and compromise its effectiveness. This motivates us to delve into the semantic vagueness problem in trusted multi-view learning.

In this paper, we propose a Consistent and Complementary-aware trusted Multi-view Learning (CCML) for this problem. First, we construct view-specific evidential Deep Neural Networks (DNNs) to learn the view-specific evidence, which represents the level of support for each category obtained from the data. In the multi-view fusion stage, we dynamically decouple the consistent and complementary evidence. The consistent evidence is derived from the consistent portions across all views, while the complementary evidence is obtained by averaging the differing portions from all views. This separation allows us to capture both the shared information and the unique aspects of each view. In the training stage, we enforce strict alignment between the opinion constructed from the consistent evidence and the ground-truth category. This is achieved by adjusting the probabilities of the true and false categories, as well as enhancing the separation between them. As for the opinion constructed from the complementary evidence, we only require it to reflect the probability of the true category, allowing for potential vagueness in the evidence. In the test stage, we aggregate the consistent and complementary evidence to make a decision. 

The main contributions of this work are summarized as follows: 1) we identify the semantic vagueness phenomenon in multi-view data, which can significantly suppress the learning of view-specific evidence in existing trusted multi-view learning methods; 2) we propose the CCML method to tackle this problem. CCML effectively addresses the negative impact of semantic vagueness through two key strategies: dynamically decoupling the consistent and complementary evidence, and allowing potential vagueness in the complementary evidence; 3) we conduct empirical comparisons between CCML and state-of-the-art trusted multi-view learning baselines on a synthetic toy dataset and six real-world datasets. The experimental results not only validate the effectiveness of the proposed dynamic decoupling strategy but also demonstrate that CCML surpasses the baseline methods in terms of accuracy and reliability.

\section{Related Work}
The proposed CCML is a new uncertainty-aware multi-view fusion methods. Therefore, in this section, we briefly review two lines of related work, multi-view fusion and uncertainty-aware deep learning, to better motivate this work.

\subsection{Multi-view Fusion}
%\textcolor{blue}{Sections 2.1 and 2.2 are related and should be summarized in a section.
%Briefly introduce multi-view learning motivation. Consistent and Complementary Principle is  ``core principle" in this area. introduce grouped.
%from earlier machine learning works to deep learning works, and LLM models.
%Problem: cannot achieve this in trusted multi-view learning.  How CCML solve this?
%}
Multi-view fusion is highly effective in a wide range of tasks, as it combines information from multiple sources or modalities \cite{fang2024representation, hu2024deep, yin2021incomplete}, use the ability to collect different information from different perspectives, leading to a more comprehensive understanding of the data. A recent comprehensive survey about multi-view fusion on low-quality data is \cite{zhang2024multimodalfusionlowqualitydata}. Based on the fusion strategy employed, existing deep multi-view fusion methods can be broadly categorized into two main pipelines: feature fusion \cite{andrew2013deep, liu2024attention} and decision fusion \cite{jillani2020multi}. Feature fusion methods aim to capture the interactions between different views at the feature level. For example, canonical correlation analysis and its variants \cite{andrew2013deep,NEURIPS2019_11b9842e} maximize the correlation of the multi-view latent representations. Matrix factorization methods \cite{xu2018deep} decode the multi-view common representation to view-specific data via basis matrices. Following this line, Xu \textit{et al.} explicitly model consistent and complementary information \cite{chao2016consensus} at the highest abstraction level by the group sparseness constraint \cite{xu2018deep}. More recently, researchers have used deep neural networks to decouple the complex consistent and complementary information at representation level \cite{xu2021multi, dong2023cross}. A major challenge of feature fusion methods is that low-quality views may adversely affect the representation of other views. Decision fusion methods \cite{roitberg2019analysis,shutova2016black,duan2018unified,morvant2014majority} mitigate this problem by integrating the decision results from different views. We follow this line and propose a new method for decoupling consistent and complementary information at the decision level.low this line and propose a new method for decoupling consistent and complementary information at the decision level.

\begin{figure*}
    \centering
    \includegraphics[width=1\textwidth]{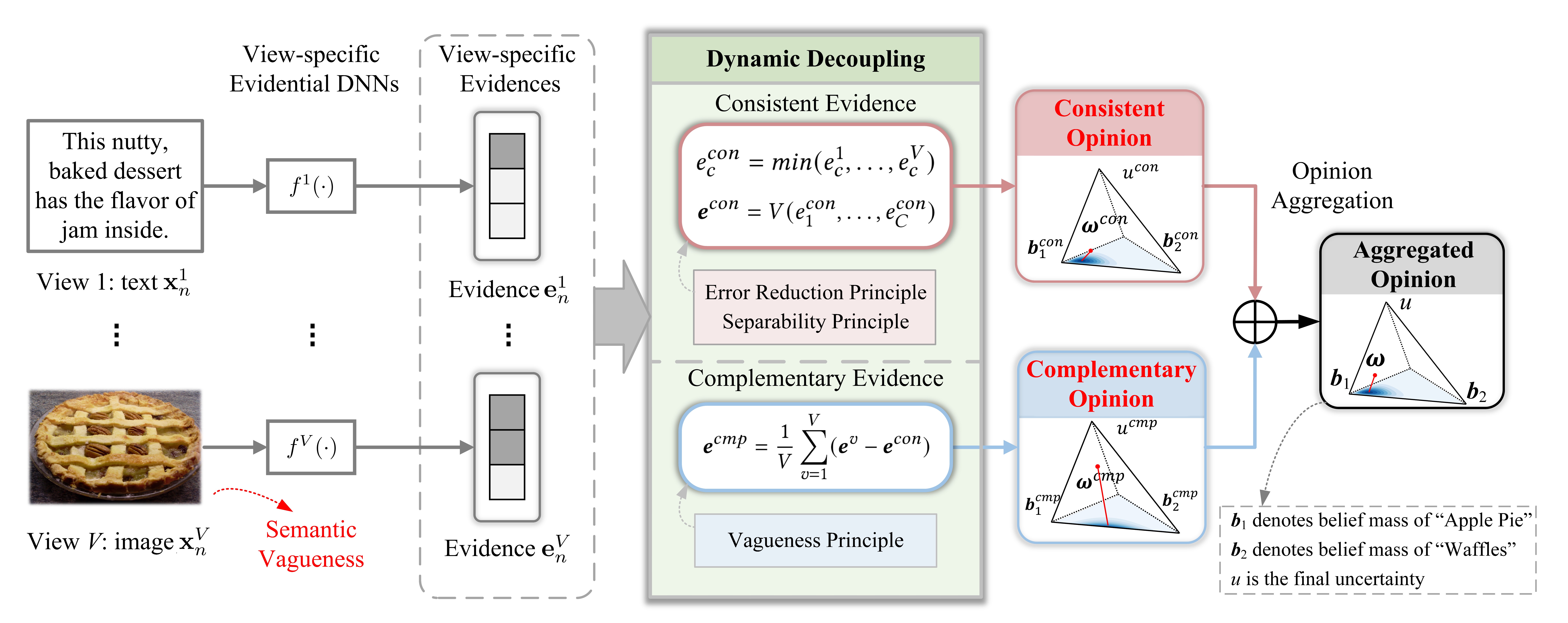}
    \caption{Illustration of CCML. CCML initially constructs view-specific evidential DNNs to acquire the view-specific evidence and subsequently dynamically decouples the consistent and complementary evidence. During training, CCML ensures precise alignment between the opinion constructed from the consistent evidence and the ground-truth category. Regarding the complementary evidence, CCML only necessitates reflecting the probability of the true category, accommodating potential evidence vagueness. During testing, CCML combines consistent and complementary evidence to reach a decision. }
    \label{fig:Schematic}
\end{figure*}

\subsection{Uncertainty-aware Deep Learning}
Traditional deep learning methods have made remarkable progress in various domains. However, they are unable to provide uncertainty estimates in predictions, which is increasingly important in real-world scenarios such as many high-risk areas. To tackle this challenge, researchers have proposed uncertainty-aware deep learning methods. One approach is Bayesian Neural Networks \cite{neal2012bayesian}, which considers the variation in results caused by variations in the data distribution as uncertainty. However, the high computational cost associated with Bayesian neural networks limits their practical applications. As a result, researchers begin to explore more efficient methods for estimating uncertainty. The Monte Carlo dropout method \cite{gal2016dropout} is one such approach. It involves using multiple instances of input data to obtain multiple prediction results, from which uncertainty measures can be calculated. Another method, Evidential Deep Learning (EDL) \cite{sensoy2018evidential} , calculates category-specific evidence and considers the lack of evidence as a source of uncertainty using a single deep neural network. Recently, Trusted Multi-view Classification (TMC) \cite{han2021tmc} extends EDL to the multi-view learning area. Following this line, researchers have proposed various evidence aggregation paradigms, including sum \cite{liu2022trusted}, (weighted) average \cite{xu2024reliable}, element-wise dot product \cite{han2022trusted}, etc. 

However, many methods often overlook the phenomenon of semantic vagueness present in real-world multi-view data \cite{zhang2024multimodalfusionlowqualitydata}, which can significantly hinder the learning of view-specific evidence. \cite{9906001} adopts a relatively static fusion strategy to solve this problem, but it cannot flexibly adapt to specific samples. Our method employs a dynamic fusion strategy, which allows our method to dynamically adjust the fusion weight based on the consistency of different samples. The proposed CCML effectively addresses the issue by allowing for potential vagueness in the complementary evidence.

\section{The Method}

In this section, we first introduce the trusted multi-view classification problem and semantic vagueness phenomenon, then present the pipeline and loss function of CCML in detail.

\subsection{Notations and Problem Definition}

In this section, we introduce the trusted multi-view classification problem and semantic vagueness phenomenon in detail. For the $C$ classification problem, considering a dataset $D=\{\{\boldsymbol{x}_n^v\}_{v=1}^V, \boldsymbol{y}_n\}_{n=1}^N$ has $N$ instances with $V$ views, where $\boldsymbol{x}_n^v$ denotes the feature vector for the $v$-th view of the $n$-th instance, and the one-hot vector $\boldsymbol{y}_n\in\{0,1\}^C$ denotes the ground-truth label of the $n$-th instance. 

The semantic vagueness phenomenon indicates that some elements of $\{\boldsymbol{x}_n^v\}$ may not well distinguish certain categories, the intuitive example is shown in figure \ref{fig:introduction}. The goal is to accurately predict $\boldsymbol{y}_n$ and provide the associated prediction uncertainties $u_n^v \in [0,1]$ which is related to the reliability of final decision.
% , the image view of the first instance is hard to differentiate between the categories ``waffles'' and ``apple pie''
% . For example, in Figure \ref{fig:introduction}, the image view of the first instance is hard to differentiate between the categories ``waffles'' and ``apple pie''.

%that cannot well distinguish certain categories (for example, due to the lack of feature information to distinguish categories $c_1$ and $c_2$, view $v$ will have similar feature vectors for instances of categories $c_1$ and $c_2$, and eventually they will not contribute to the classification of $c_1$ and $c_2$), which can be called vague information views. The goal is to accurately predict the final result from $D$ and provide the attached prediction uncertainties $u_n^v \in [0,1]$ to measure the decision reliability.

\subsection{CCML Pipeline}

As shown in Figure \ref{fig:Schematic}, the overall architecture is a decision-level fusion pipeline, which consists of the view-specific evidence learning stage and the evidential multi-view fusion stage. The view-specific DNNs $\{ f^v( \cdot ) \}_{v=1}^V$ learn the view-specific evidence, which indicates the level of support for each category based on the data. In the fusion stage, we dynamically decouple the consistent evidence ($\boldsymbol{e}^{con}$) and complementary evidence ($\boldsymbol{e}^{cmp}$). $\boldsymbol{e}^{con}$ is extracted from the consistent portions across all views, while $\boldsymbol{e}^{cmp}$ is obtained by averaging the differing portions across all views. In the training stage, we establish different principles for $\boldsymbol{e}^{con}$ and $\boldsymbol{e}^{cmp}$, respectively. For the testing stage, we aggregate $\boldsymbol{e}^{con}$ and $\boldsymbol{e}^{cmp}$ to make a decision.

%Many existing uncertainty-aware multi-view learning methods cannot solve the VIMC problem very well. Therefore, we have developed a method called Consistent and Complementary Multi-view Learning (CCML) which effectively utilizes highly complementary information between different views to solve the VIMC problem and also demonstrates good performance on normal real-world datasets. The CCML process is shown in the Figure \ref{fig:Schematic}.

\subsubsection{View-specific Evidence Learning}

%Many existing multi-view learning methods with traditional neural classification networks usually use the softmax layer as the standard output to solve multi-classification problems. However, these methods only obtain the class probabilities but ignore the reliability of the results \cite{hendrycks2016baseline}, because the softmax score provides a single-point estimation of the predictive distribution, it results in overconfident outputs even in the case of incorrect predictions. To solve the above problems and accurately predict uncertainty, we adopt EDL \cite{sensoy2018evidential} by replacing the softmax layer with ReLU to obtain non-negative evidence. 

Many conventional multi-view learning methods utilize softmax layers to produce standard outputs to address multi-classification problems in neural classification networks. However, these methods only provide class probabilities without considering the reliability of the results \cite{hendrycks2016baseline}. Due to the single-point estimation paradigm of softmax scores, they tend to produce overconfident outputs, even when the predictions are incorrect. To address these issues and achieve accurate uncertainty prediction, we employ EDL \cite{sensoy2018evidential} by replacing the softmax layer with a ReLU activation to obtain non-negative evidence. 

%To realize multi-view fusion, in this section, we will introduce view-specific evidence deep learning, which is used to generate view-specific evidence and estimate the classification uncertainty. For VIMC problems, this stage can guarantee that the evidence generated by vague views has the support amount for multiple categories. The indistinguishable categories that have the true categories in these vague views produce similar and significantly higher amounts of evidence than other categories that can be identified as errors.

We also introduce the subjective logic \cite{josang2016subjective} framework to form opinions which is important in CCML. In this framework, the parameter $\bm{\alpha}$ of the Dirichlet distribution $Dir(\bm{p}|\bm{\alpha})$ is associated with the belief distribution in the framework of evidence theory, where $\bm{p}$ is a simplex representing the probability of class assignment. We collect evidence $\{ \boldsymbol{\mathit{e}}_{n}^{v}  \}$, using view-specific evidential DNNs $\{ f^v (\cdot) \}_{v=1}^V$. The corresponding Dirichlet distribution parameters are $\bm{\alpha}^v = \bm{e}^v+1=[\alpha_1^v,\cdots,\alpha_C^v]^T$. After obtaining the distribution parameters, we can calculate the subjective opinion, $\boldsymbol{\omega}^v=({\bm{b}^v,u^v})$ of the view including the quality of beliefs $\bm{b}^v$ and the quality measure of uncertainty $u$, where $\bm{b}^v=(\bm{\alpha}^v - 1) / S^v=\bm{e}^v / S^v$, $u^v=K / S^v$, and $S^v=\sum_{k=1}^K\alpha_k^v$ is the Dirichlet intensity.

After training evidential DNNs to learn view-specific evidence, we observe that the evidence produced by vagueness views contains support for multiple categories. Among these categories, the true categories would generate relatively higher amounts of evidence than other categories. Therefore, we cannot require only the ground-truth category to have large evidence. We utilize this property in the subsequent evidence fusion stage.

\subsubsection{Consistent and Complementary-aware Multi-view Fusion}

%\textcolor{red}{A paragraph point the motivation of our method. similar as introduction and related work. A paragraph summarizes the principles of the fusion. different principles for the Consistent and Complementary evidences.
%}

%Existing trusted multi-view learning methods are limited in their performance when facing semantic vagueness phenomenon. The reason is that in the process of generating view-specific evidence, the indistinguishable error categories in the vague views generate the same amount of evidence as the indistinguishable correct categories, and many methods fail to differentiate between the same amounts of evidence for the correct and error categories. Therefore, this motivates us to solve this problem by using the complementary relation between different views, extracting evidence for categories with higher evidence in all views, and reducing the amount of probable correct categories through complementary discriminative abilities between different views.

Existing trusted multi-view learning methods face limitations when addressing the phenomenon of semantic vagueness. This is primarily because, during the generation of view-specific evidence from network, the indistinguishable error categories in vague views produce a similar amount of evidence as the indistinguishable categories. Many methods struggle to distinguish between equal amounts of evidence for the correct and incorrect categories. As a result, we are motivated to tackle this problem by the CCML which decoupling the consistent and complementary evidence and allowing for potential vagueness in the complementary evidence.

For views $v_i$ and $v_j$, the view-specific deep learning network of view $v_i$ will produce similar amounts of evidence for classes $c_1$ and $c_2$, while the network of view $v_j$ will produce similar amounts of evidence for classes $c_2$ and $c_3$ when classifying instances of class $c_2$, Effective classification should be achieved through complementary information from view $v_i$ and $v_j$ respectively ($v_i$ determines that the instance is one of $c_1$ or $c_2$, and $v_j$ determines that the instance is one of $c_2$ or $c_3$, the final classification result should be $c_2$ utilizing the complementary information of $v_i$ and $v_j$). This also demonstrates the basic idea of the CCML method, which is to fully utilize consistent evidence across different views.

%\textcolor{blue}{Dynamically Decoupling the Consistent and Complementary Evidences. Should point out why do this, the advantage of Dynamically compared to fixed relations. (IEEE TII, EMDL)}

Therefore, we need to decouple view-specific evidence of different views into consistent and complementary evidence. One previous work \cite{9906001} addresses the semantic vagueness phenomenon by setting fixed relation degradation layers for semantic vagueness categories. However, the extent of semantic vagueness varies among different instances, as reflected by the difference in consistency and complementarity. It is necessary to use dynamic decoupling strategies according to the differences of consistency and complementarity in the classification tasks, instead of using fixed relations. For this purpose, we propose a dynamic consistent and complementary evidence decoupling strategy.

\begin{definition}{(Consistent and Complementary Evidences).}
    For the view-specific evidences $\{ \boldsymbol{e}^v=(e_1^v,\ldots,e^v_C) \}_{v=1}^V$. The consistent evidence $\boldsymbol{e}^{con}$ and complementary evidence $\boldsymbol{e}^{cmp}$ are defined as:
    \begin{subequations}
    \begin{align}
        & \boldsymbol{e}^{con} = V (e_1^{con},\ldots,e_C^{con}), \\
        & e_c^{con} = min(e_c^1,\ldots,e_c^V), c=1,\ldots,C, \\
        & \boldsymbol{e}^{cmp} = \frac{1}{V}\sum_{v=1}^V(\boldsymbol{e}^v-\boldsymbol{e}^{con}),
    \end{align}
    \end{subequations}
    where $\boldsymbol{e}^{con}$ denotes the minimum evidence of each class supported by each view. $\boldsymbol{e}^{cmp}$ denotes the complementary evidence.
\end{definition}

The consistent evidence $\boldsymbol{e}^{con}$ is obtained by aggregating the consistent portions from all views, while the complementary evidence $\boldsymbol{e}^{cmp}$ is calculated as the average of the differing portions across all views. This distinction is made because the complementary evidence of all views $ \{ \boldsymbol{e}^v-\boldsymbol{e}^{con} \}_{v=1}^V $ capture the varying information between different views, encompassing both complementarity and conflict. It is important to note that this information does not always enhance the accuracy of results; instead, it can introduce additional uncertainty. Consequently, retaining the entire set of complementary evidence is not advisable. Instead, a more suitable approach is to preserve it by taking the average \cite{xu2024reliable}. By employing this strategy, we dynamically separate the consistent and complementary evidence from the view-specific evidence. Finally, we aggregate the two pieces of evidence on average to form the final opinion, which is used in the test stage. In the following section, we will outline how our deep learning networks are trained using distinct principles for handling consistent and complementary evidence.

% \textcolor{blue}{Define $\boldsymbol{e}^{cmp} = \frac{1}{V} \sum_{v=1}^V(\boldsymbol{e}^v-\boldsymbol{e}^{con})$. explain the reason (refer as conflictive fusion, the overall evidence may small as the view-specific) }

\subsection{Loss Function}
% \textcolor{blue}{Summarize xxx. The goal is xxx.  }
%In this section, we will introduce the training DNN to obtain the fused multi-view opinion. The goal is to process the decoupled consistent and complementary evidence separately based on the view-specific evidence generated from all views. The consistent evidence, as a common part of all views, should closely approximate the true value, while the complementary evidence, serving as a supplement to the consistent evidence, allows more relaxed constraints. We now describe different components of our loss function.

In this section, we will present the training process of CCML. The objective is to deal with view-specific, consistent, and complementary evidence separately. Specifically, the consistent evidence should closely approximate the true label, while the complementary evidence serves as a supplementary component, allowing for more relaxed constraints. We will elaborate on these components below.
 
\subsubsection{View-specific Loss Function}

The evidential DNNs are obtained by converting the softmax layer of traditional DNNs into ReLU. Therefore, we obtain the non-negative outputs as evidence. We introduce an adjusted cross-entropy loss function to ensure that all views can generate appropriate non-negative view-specific evidence for classification:
\begin{align}
    \mathcal{L}_{ace}(\boldsymbol{\alpha}_n) =   & \int\left[\sum_{j=1}^C-y_{nj}logp_{nj}\right]\frac{1}{B(\boldsymbol{\alpha_n})}\sum_{j=1}^Cp_{nj}^{\alpha_{nj}-1}d\boldsymbol{p}_n \nonumber \\ 
    =           & \sum_{j=1}^Cy_{nj}(\psi(S_n)-\psi(\alpha_{nj})),
\end{align}
where $\psi(\cdot)$ is the digamma function. And to grarantee that the evidence generated by the incorrect labels is lower, we introduce the Kullback-Leibler (KL) divergence into the loss function:

\begin{align}
    \mathcal{L}_{KL}(\boldsymbol{\alpha}_n) = & \lambda_t KL[D(\boldsymbol{p}_n|\widetilde{\boldsymbol{\alpha}}_n)\,||\,D(\boldsymbol{p}_n|\boldsymbol{1})] \nonumber \\
    = & log\left(\frac{\Gamma(\sum_{c=1}^C\widetilde{\alpha}_{nc})}{\Gamma(K)\prod_{c=1}^C\Gamma(\widetilde{\alpha}_{nc})}\right) \nonumber \\
    + & \sum_{c=1}^C(\widetilde{\alpha}_{nc}-1)\left[  \psi(\widetilde{\alpha}_{nc})-\psi(\sum_{j=1}^C\widetilde{\alpha}_{nj}) \right],
\end{align}
where $D(\boldsymbol{p}_n|\boldsymbol{1})$ is the uniform Dirichlet distribution, $\widetilde{\boldsymbol{\alpha}}_n = \boldsymbol{y}_n+(\boldsymbol{1}-\boldsymbol{y}_n)\odot\boldsymbol{\alpha}_n$ is the Dirichlet distribution parameter after removing the evidence of the ground-truth category from the original parameter $\boldsymbol{\alpha}_n$. $\lambda_t = min(1, t/T) \in [0,1]$ is the annealing coefficient, acting as the balance factor in the training process. As the training process progresses, $\lambda_t$ continues to increase, enhancing the influence of KL divergence accordingly, to prevent premature convergence of misclassified instances to the uniform distribution.

The view-specific loss function of $\boldsymbol{x}_n^v$ is defined as:
\begin{equation}
    \mathcal{L}_{vs}(\boldsymbol{\alpha}_n^v) = \mathcal{L}_{ace}(\boldsymbol{\alpha}_n^v)+\mathcal{L}_{KL}(\boldsymbol{\alpha}_n^v),
\end{equation}
where $\boldsymbol{\alpha}_n^v = \boldsymbol{e}_n^v + 1$ is the parameters of the corresponding Dirichlet distribution, $\boldsymbol{e}_n^v = f^v(\boldsymbol{x}_n^v)$ represent the evidence vector predicted by the view-specific neural network.

% Here we do not add the KL loss term to each view's Dirichlet parameter $\boldsymbol{\alpha}_n^v$ separately, but only add it to the aggregated Dirichlet parameter $\boldsymbol{\alpha}_n$. This ensures that all single-view networks provide sufficient evidence for the VIMC problem without suppressing them due to the ambiguity of views. 

\subsubsection{Consistent Loss Function}

%\textcolor{blue}{Overview - 2 criteria: separability across categories, complying with the label (the name should be considered repeatedly)}

The consistent evidence is obtained by aggregating the consistent portions of all views. Consequently, we impose a stringent alignment between the opinion constructed from the consistent evidence and the ground-truth category. To accomplish this, we adjust the probabilities assigned to the true and false classes while simultaneously enhancing the distinction between them. To achieve this objective, we introduce two principles: the Error Reduction Principle and the Separability Principle.

% \begin{align}
%     \mathcal{L}_{KL}(\boldsymbol{\alpha}_n) = & \lambda_t KL[D(\boldsymbol{p}_n|\widetilde{\boldsymbol{\alpha}}_n)\,||\,D(\boldsymbol{p}_n|\boldsymbol{1})] \nonumber \\
%     = & log\left(\frac{\Gamma(\sum_{c=1}^C\widetilde{\alpha}_{nc})}{\Gamma(K)\prod_{c=1}^C\Gamma(\widetilde{\alpha}_{nc})}\right) \nonumber \\
%     + & \sum_{c=1}^C(\widetilde{\alpha}_{nc}-1)\left[  \psi(\widetilde{\alpha}_{nc})-\psi(\sum_{j=1}^C\widetilde{\alpha}_{nj}) \right],
% \end{align}
% where $D(\boldsymbol{p}_n|\boldsymbol{1})$ is the uniform Dirichlet distribution, $\widetilde{\boldsymbol{\alpha}}_n = \boldsymbol{y}_n+(\boldsymbol{1}-\boldsymbol{y}_n)\odot\boldsymbol{\alpha}_n$ is the Dirichlet distribution parameter after removing the evidence of the ground-truth category from the original parameter $\boldsymbol{\alpha}_n$. $\lambda_t = min(1, t/T) \in [0,1]$ is annealing coefficient, acting as the balance factor. As the training process progresses, $\lambda_t$ continues to increase, enhancing the influence of KL divergence, to prevent premature convergence of misclassified instances to the uniform distribution.

\paragraph{Separability Principle}
% \textcolor{blue}{Definition 1 separability principle. From regularization to equivalent self-enhancement}
The Separability Principle emphasizes the importance of creating a significant distinction between the evidence supporting different categories during the classification process. This principle allows the classifier to more accurately distinguish between different categories, especially semantic vagueness categories. For instance, consider the belief masses of opinions $\boldsymbol{b}^1=(0.4, 0.4)$ and $\boldsymbol{b}^2=(0.3,0)$. The total amount of belief mass in $\boldsymbol{b}^2$ is more than $\boldsymbol{b}^1$. However, the greater degree of separation of $\boldsymbol{b}^2$ makes it more contribution to the classification result than $\boldsymbol{b}^1$. To enhance classification performance, we need to increase the degree of separation during the training process. Therefore, we quantify the degree of separation of the belief masses supporting different classes in opinion using the separation degree.

\begin{definition}{(Separation Degree).}
    For the subjective opinion $\boldsymbol{\omega} = (\boldsymbol{b}, u, \boldsymbol{a})$, where $\boldsymbol{b} = (b_1,\ldots,b_C)$, the separation degree can be defined as:
    \begin{equation}
        SD(\boldsymbol{b}) = \sum_{i=1}^C\sum_{i\neq j}^C|b_i-b_j|.
    \end{equation}
\end{definition}

Our goal is to increase the separation degree of consistent opinions. There are two approaches to increase it: the first approach involves adding a constraint to maximize the degree of separation. However, this approach may lead to an increase in the total amount of evidence across all categories. This is contrary to our original intention of controlling uncertainty based on the total amount of evidence. Therefore, we design the following method.

We first convert consistent evidence into consistent opinions $\boldsymbol{\omega}^{con} = (\boldsymbol{b}^{con}, u^{con}, \boldsymbol{a}^{con})$ and adjust it to obtain the final consistent opinion $\widetilde{\boldsymbol{\omega}^{con}} = ( \widetilde{\boldsymbol{b}^{con}}, u^{con},\boldsymbol{a}^{con})$:
\begin{equation}
    \widetilde{\boldsymbol{b}^{con}} =
    \left\{
        \begin{array}{cc}
             \frac{\sum_{c=1}^Cb_c^{con}}{\sum_{c=1}^C\widetilde{b_c^{con}}}(\widetilde{b_1^{con}},\ldots, \widetilde{b_C^{con}}), & \sum_{c=1}^C\widetilde{b_c^{con}} \neq 0, 
            \\0, & otherwise, 
        \end{array}
    \right.
\end{equation}
where $\widetilde{b_c^{con}} = pow(b_c^{con}, \beta) \in [0,1]$ is the belief mass supporting classes $c$, $\beta$ is a hyper-parameter which is bigger than $1$. For $\boldsymbol{a}^{con}$, we follow related works to set each $a_c^{con}$ as $1/C$ without prior evidence. By the power operation $pow(b_c^{con}, \beta)$, the separation degree would increase. The reason behind this is the increasing disparity in the confidence mass that supports each category. We also demonstrate theoretically that this approach can increase the separation degree, which is elaborated in the appendix. Therefore, we only need this simple operation to achieve the increase of separation degree without changing the total amount of evidence.

\paragraph{Error Reduction Principle}
% \textcolor{blue}{Definition 2 xxx principle }

The error reduction principle highlights that during the training process, the evidence generated for incorrect categories may inadvertently increase due to the inherently limited availability of counterexamples. This misleading evidence has the potential to introduce challenges to the classification process. Therefore, we also use the KL divergence to actively reduce evidence for incorrect labels categories.

Under the guidance of the above two principles, we can derive the consistent loss function:
\begin{equation}
    \mathcal{L}_{con}(\boldsymbol{\alpha}_n^{con}) = \mathcal{L}_{ace}(\widetilde{\boldsymbol{\alpha}^{con}}) + \eta \mathcal{L}_{KL}(\widetilde{\boldsymbol{\alpha}^{con}}).
\end{equation}
where $\widetilde{\boldsymbol{\alpha}^{con}}$ represents the corresponding parameters of the final Dirichlet distribution of the consistent opinion. The consistent loss function can refine the consistent evidence. After increasing the separation degree, the cross-entropy loss function minimizes the model's deviation from the true label, and the KL loss function reduces the evidence for incorrect categories. The combination of these two loss functions can maximize the impact of consistent evidence on the model training process.

\subsubsection{Complementary Loss Function}

\paragraph{Vagueness Principle}
Complementary evidence usually represents complementary or even conflicting information between different views. Its reliability is generally lower than that of the consistent evidence. As a supplement to the consistent evidence, the complementary evidence need not necessarily enhance the degree of separation or reduce false labels like consistent evidence. It accounts for potential vagueness in multi-view data. Therefore, it is only required to reflect the probability of the true category.

According to this principle, we define the complementary loss function as follows:
\begin{equation}
    \mathcal{L}_{cmp}(\boldsymbol{\alpha}_n^{cmp}) = \sum_{j=1}^Cy_{nj}(\psi(S_n^{cmp})-\psi(\alpha_{nj}^{cmp})),
\end{equation}
where $\boldsymbol{\alpha}_n^{cmp} = \boldsymbol{e}_n^{cmp} + 1$ is the corresponding Dirichlet parameter of the complementary evidence.

\subsubsection{Joint Loss}

By synthesizing the above objectives, the overall loss function for a specific instance $\{\boldsymbol{x}_n^v\}_{v=1}^V$ is formulated as:
\begin{equation}
    \mathcal{L} = \sum_{v=1}^V \mathcal{L}_{vs}(\boldsymbol{\alpha}_n^v) + \delta \mathcal{L}_{con}(\boldsymbol{\alpha}_n^{con}) + \gamma \mathcal{L}_{cmp}(\boldsymbol{\alpha}_n^{cmp}),
\end{equation}
where $\delta,\gamma > 0$ are hyper-parameters that we can adjust.

\section{Experiment}
In this section, we show the empirical results of CCML in making trusted decisions for multi-view inputs. We first apply CCML on a synthetic toy example to investigate its performance in solving the semantic vagueness problem, then we evaluate CCML on six real-world multi-view datasets, compare it with existing multi-view learning methods and conduct other analysis experiments.

\subsection{A Toy Example}

The major advantage of CCML compared with pioneer uncertainty-aware methods is the ability to perceive consistent and complementary information between different views. Therefore, we conducted a set of comparative experiments with TMC in the Toy Dataset to investigate the effectiveness of CCML in solving semantic vagueness questions and to explicitly achieve higher accuracy results. 

We set view 1 cannot distinguish categories $c_2$ and $c_3$ and view 2 cannot distinguish categories $c_1$ and $c_2$, respectively. Specifically, the toy dataset consists of 2 views, each with $1200$ data instances $\{\boldsymbol{x}_{n}^v\}_{n=1}^{1200}$ belonging to 3 categories, $c_1$, $c_2$, and $c_3$, with $400$ data instances in each category. The underlying latent space has $9$ dimensions, with three for each category. The first 3 dimensions and the last 3 dimensions are private to categories $c_1$ and $c_3$ respectively, and the middle dimensions are a shared dimension for $c_2$ and $c_3$. Each element of $\{\boldsymbol{v}_{n}^v\}_{n=1}^{1200}$ is the sum of a number sampled from a gamma-distributed $\Gamma(1,0.9)$, the noise sampled from Gaussian distribution $N(0,0.1)$, and a consistent term of $0.5$. We randomly generated $12 \times 9$ basis matrices $\boldsymbol{W}^v$ for the two views, with elements drawn from a uniform distribution $U(0.4,1)$, We randomly set 30 percent of the elements to be zero to simulate the real-world multi-view mapping pattern. Then we generated a noise matrix $\boldsymbol{Z}^v$, and the elements of $\boldsymbol{Z}^v$ drawn from by the Gaussian distributions $N(0,0.5)$ and $N(0,1)$, respectively. We use the equation $\boldsymbol{x}_{n}^v = \boldsymbol{W}^v\boldsymbol{v}_n^v+\boldsymbol{z}^v$ to generate data instances.

\begin{figure}
    \centering
    \subfloat[view 1]
      {
          \label{fig:toy_instances1}\includegraphics[width=0.225\textwidth]{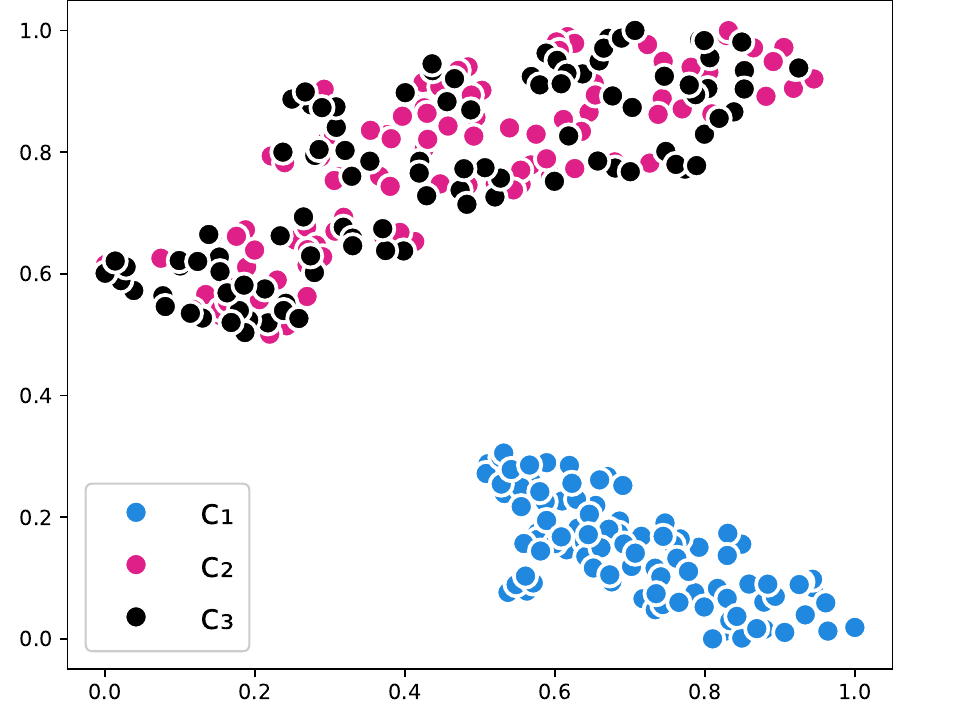}
      }
      \subfloat[view 2]
      {
          \label{fig:toy_instances2}\includegraphics[width=0.225\textwidth]{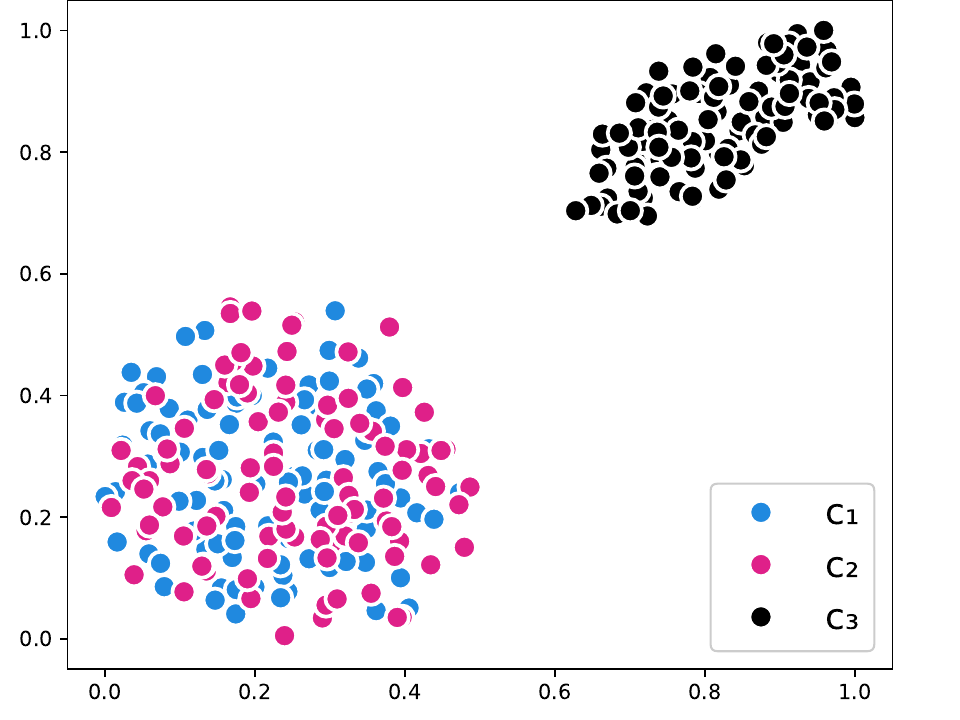}
      }
    \caption{\label{fig:toy_instances} Visualization of data instances in the toy dataset.}
\end{figure}

In the Toy Dataset, we set the last 3 columns of $\boldsymbol{W}^1$ to be 0 and the first 6 columns of $\boldsymbol{W}^2$ to be 0, respectively. Therefore, the data instances in view 1 cannot distinguish between categories $c_2$ and $c_3$, and the data instances in view 2 cannot distinguish between categories $c_1$ and $c_2$. The Toy Dataset represents a strong complementarity between the two perspectives. For a more intuitive explanation, we use t-SNE to visualize the multi-view data instances of the Toy Dataset, as shown in Figure \ref{fig:toy_instances}.

\begin{table}
    \centering
    \caption{Classification accuracy (\%) and uncertainty on the toy dataset.}
    \begin{tabular}{ccc}
        \toprule
          & Accuracy & Uncertainty  \\
        \midrule
        TMC & $94.73\pm 0.25$ & $0.58$ \\
        \midrule
        CCML & $98.07\pm 0.28$ & $0.23$  \\
        \bottomrule 
    \end{tabular}
    \label{tab:toy}
\end{table}

\begin{figure}
    \centering
    \subfloat[TMC]
      {
          \label{fig:toy_evidence1}\includegraphics[width=0.225\textwidth]{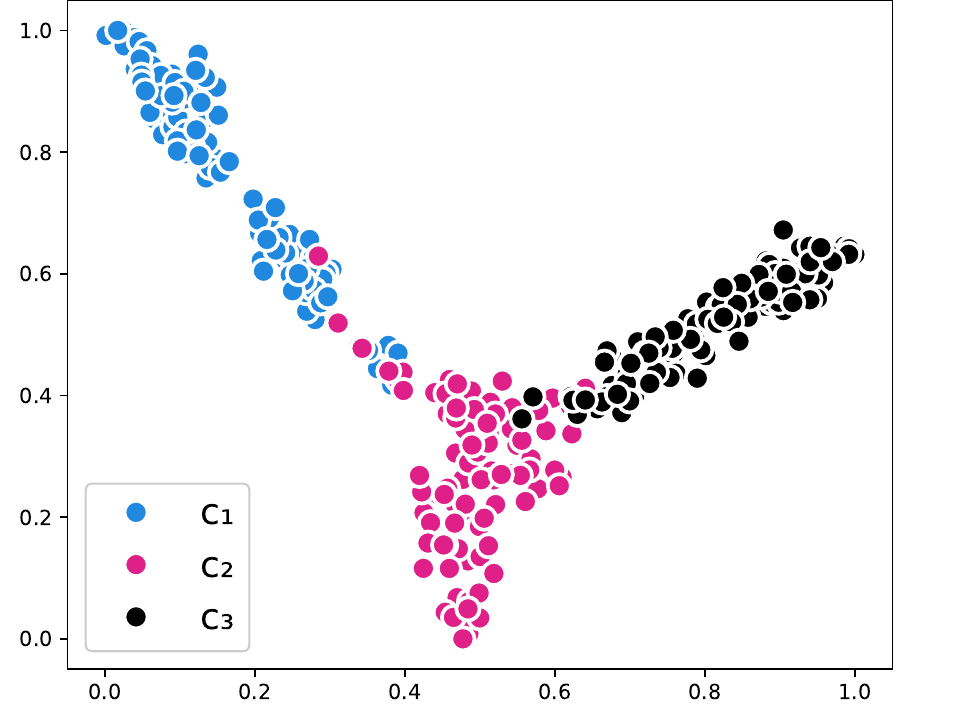}
      }
      \subfloat[CCML]
      {
          \label{fig:toy_evidence2}\includegraphics[width=0.225\textwidth]{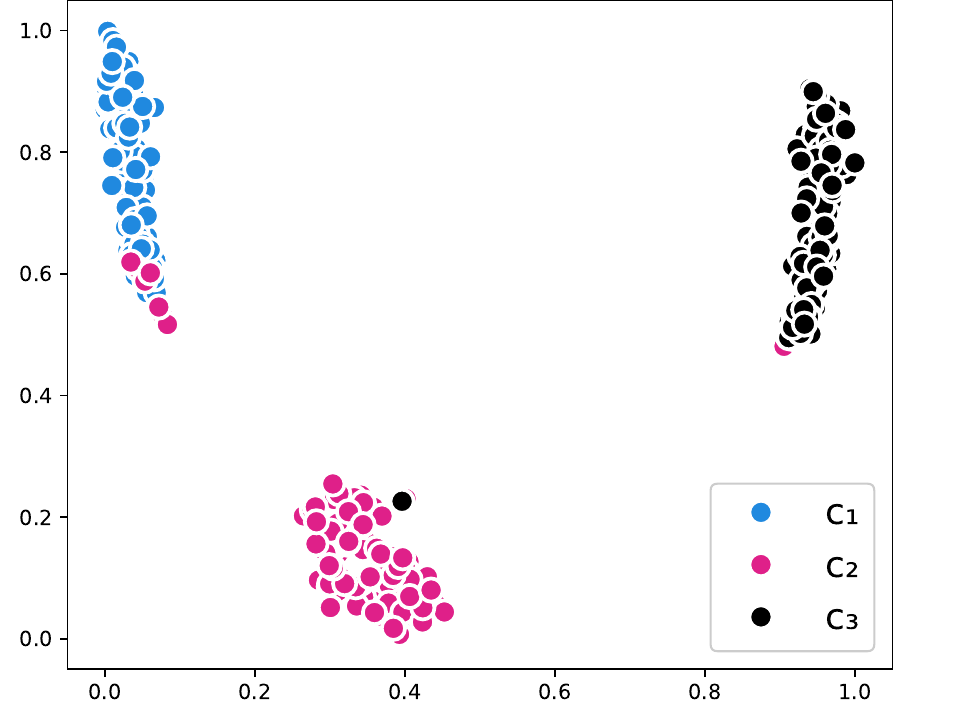}
      }
    % \caption{\label{fig:toy_evidence}The evidences of instances of $c_2$ on the toy dataset.}
    \caption{\label{fig:toy_evidence}Visualization of the evidences (outputs of TMC and CCML on the toy dataset.}
\end{figure}

\begin{table*}
    \caption{Classification accuracy (\%) on different datasets.}
    \centering
    \begin{tabular}{c|ccccccc|c}
        \toprule
        Data & EDL & IEDL & DCCAE & DCP & CAML & ETMC & RCML & CCML \\
        \midrule
        Handwritten & $97.00\pm 0.16$ & $98.45\pm0.43$ & $97.05\pm 0.24$ & $88.10\pm1.09$ & $98.10\pm 0.12$ & $98.32\pm 0.22$ & $\underline{98.70\pm 0.19}$ & $\boldsymbol{98.75\pm 0.27}$ \\
        Scene15 & $60.60\pm 0.13$ & $65.40\pm1.70$ & $64.26\pm 0.42$ & $72.08\pm1.65$ & $70.17\pm 0.13$ & $66.87\pm 0.29$ & $\underline{71.28\pm0.32}$ & $\boldsymbol{74.76\pm 0.85}$  \\
        CUB & $89.51\pm 0.24$ & $92.67\pm2.35$ & $85.39\pm 1.36$ & $93.00\pm2.33$ & $\underline{94.33\pm 0.73}$ & $91.05\pm 0.63$ & $93.28\pm2.75$ & $\boldsymbol{94.58\pm 1.30}$ \\
        LandUse & $47.10\pm1.71$ & $49.15\pm 0.28$ & $50.42\pm 0.26$ & $53.43\pm3.67$ & $\underline{58.18\pm1.21}$ & $52.06\pm 0.71$ & $53.55\pm 0.33$ & $\boldsymbol{58.70\pm 1.75}$\\
        PIE & $87.99\pm 0.56$ & $90.85\pm3.31$ & $81.96\pm 1.04$ & $87.24\pm2.48$ & $93.38\pm0.80$ & $93.82\pm 0.82$ & $\underline{93.89\pm2.46}$ & $\boldsymbol{94.56\pm 1.83}$\\
        Colored-MNIST & $38.41\pm0.43$ & $44.83\pm3.23$ & $40.35\pm0.67$ & $\underline{87.15\pm.0.58}$ & $80.26\pm 0.39$ & $83.76\pm 1.27$ & $42.11\pm2.01$ & $\boldsymbol{91.54\pm1.48}$\\
        \bottomrule 
    \end{tabular}
    \label{tab:results}
\end{table*}

% In addition, to obtain a better understanding of the advantages of CCML over TMC in addressing semantic vagueness problems, we have visualized the evidence produced by CCML and TMC for instances in category $c_2$, as illustrated in Figure \ref{fig:toy_evidence}. 

In addition, to obtain a better and intuitive understanding of the advantages of CCML over TMC in addressing semantic vagueness problems, we also conduct t-SNE visualization on the test samples by evidences of different categories during testing process of two methods, the result as illustrated in Figure \ref{fig:toy_evidence}. 

Based on the experimental results, we can obtain the following conclusions: (1) The accuracy rate of CCML on the Toy Dataset is $98.07\%$, which is significantly higher than the accuracy rate of TMC on the same dataset, which is $94.73\%$. This indicates that CCML demonstrates superior performance in addressing semantic vagueness problems. (2) Upon observation in Figure \ref{fig:toy_evidence}, it is evident that when facing semantic vagueness problems, our CCML method achieves better separation performance for different categories compared to the TMC. This further verifies the effectiveness of our model in addressing the semantic vagueness problem. 

% The evidence generated by a single view in the TMC method is subdued due to the inherent ambiguity of the view. Consequently, this results in insufficient evidence. The insufficiency arises because vague views fail to distinguish between certain categories, resulting in decision conflicts within the TMC framework, and ultimately, an inadequate evidence supply for both categories. In contrast, CCML can effectively generate sufficient evidence from each view and effectively leverage the information from vague views through the consistent evidence component. As a result, CCML constructs accurate and effective evidence supporting the $c_2$ category. 

\subsection{Experiment on Real-world Datasets}

\subsubsection{Experimental Setup}
\paragraph{Datasets.} \textbf{HandWritten}\footnote{\url{https://archive.ics.uci.edu/dataset/72/multiple+features}} comprises 2000 instances of handwritten numerals ranging from `0' to `9', with 200 patterns per class. It is represented using six feature sets. \textbf{Scene15}\footnote{\url{https://figshare.com/articles/dataset/15-Scene_Image_Dataset/7007177/1}} includes 4485 images from 15 indoor and outdoor scene categories. We extract three types of features HOG, LBP, and GIST. \textbf{CUB} \cite{wah2011caltech} consists of 11788 instances associated with text descriptions of 200 different categories of birds, we focus on the first 10 categories and extract image features using GoogleNet and corresponding text features using doc2vec. \textbf{LandUse} \cite{yang2010bag} comprises 2100 images from 21 classes. We extract HOG and SIFT features as two views. \textbf{PIE}\footnote{\url{http://www.cs.cmu.edu/afs/cs/project/PIE/MultiPie/Multi-Pie/Home.html}} contains 680 facial instances belonging to 68 classes. We extract intensity, LBP, and Gabor as 3 views. \textbf{Colored-MNIST}\footnote{\url{https://www.kaggle.com/datasets/youssifhisham/colored-mnist-dataset/}} includes 18835 instances of numerals with RGB coloured backgrounds which consist of three colours (red, green, blue) for each number. We focus on the 1200 instances and extract RGB and HOG features as two views. 

\paragraph{Compared Methods.}(1) \textbf{Single-view uncertainty aware methods} contain: EDL (Evidential Deep Learning) \cite{sensoy2018evidential} and IEDL \cite{deng2023uncertainty} which is the SOTA method that combines Fisher's information matrix. (2) \textbf{Multi-view feature fusion methods} contain: DCCAE (Deep Canonically Correlated AutoEncoders) \cite{wang2015deep} is the classical method, which employs autoencoders to seek a common representation, DCP (Dual Contrastive Prediction) \cite{lin2022dual} is the SOTA method that obtains a consistent representation. (3) \textbf{Multi-view decision fusion methods} contain: CALM (Enhanced Encoding and Confidence evaluating framework) \cite{zhou2023calm} takes advantage of cross-view consistency and diversity to improve the efficacy of the learned latent representation, ETMC (Enhanced Trusted Multi-view Classification) \cite{han2022trusted}, addresses the uncertainty estimation problem and produces reliable classification results. RCML (Reliable Conflictive Multiview Learning) \cite{xu2024reliable} is the SOTA method that proposed a fusion strategy for solving multi-view conflictive problems.

\paragraph{Implementation Details.}We briefly introduce the details of the experiment. We utilize fully connected networks with a ReLU layer to extract view-specific evidence. The Adam optimizer \cite{kingma2014adam} is used to train the network, where L2-norm regularization is set to $1e^{-5}$. We employ 5-fold cross-validation to select the learning rate from the options of $3e^{-3}$, $1e^{-3}$, $3e^{-4}$, $1e^{-4}$. In all datasets, $20\%$ of the instances are allocated as the test set. The average performance is reported by running each test case five times.

\subsubsection{Performance Comparison}

We compare CCML with the other classification methods, and the results are shown in Table \ref{tab:results}. We can obtain that: (1) multi-view methods generally outperform single-view methods, which illustrates the necessity of using multiple views in classification tasks. (2) There is a large difference in accuracy among different methods on the Colored-MNIST dataset, demonstrating significant variation in the ability of different methods to solve the semantic vagueness problem. (3) For the majority of real-world datasets, CCML shows performance comparable to state-of-the-art methods and has outstanding performance on Colored-MNIST datasets. (4) This result indicates that CCML significantly improves the ability to handle the semantic vagueness phenomenon while ensuring good performance on general classification tasks. The reason would be attributed to the consistent and complementary dynamic decoupling method, we would further verify this in the next analysis and other experiments.

\begin{table*}
    \centering
    \caption{Comparison of CCML and variants on the CUB dataset.}
    \begin{tabular}{c|ccccccc}
        \toprule
        Data & \multicolumn{7}{c}{CUB} \\
        \midrule
        Consistent Evidence & $\checkmark$ & - & $\checkmark$ & $\checkmark$ & $\checkmark$ & $\checkmark$ & $\checkmark$\\
        Complementary Evidence & - & $\checkmark$ & $\checkmark$ & $\checkmark$ & $\checkmark$ & $\checkmark$ & $\checkmark$ \\
        Aggregate Strategy & - & - & Average & Accumulate & Average & Accumulate & CCML \\
        Separation & - & - & - & - & $\checkmark$ & $\checkmark$ & $\checkmark$ \\
        \midrule
        ACC(\%) & $90.08\pm 0.80$ & $74.46\pm 3.24$ & $90.07\pm 0.32$ & $91.54\pm 0.83$ & $92.51\pm 0.32$ & $92.19\pm0.64$ & $94.58\pm 0.24$ \\
        \bottomrule
    \end{tabular}
    \label{tab:CC}
\end{table*}

\begin{table}
    \centering
    \caption{The ablation study on the Colored-MNIST dataset.}
    \begin{tabular}{ccc|c}
        \toprule
         \multicolumn{3}{c}{Modules} & ACC(\%) \\
         \midrule
         Decoupling & KL-divergence & Separation & Colored-MNIST \\
         \midrule
         - & - & - & $57.50\pm2.77$ \\
         $\checkmark$ & - & - & $80.37\pm2.53$  \\
         $\checkmark$ & $\checkmark$ & - & $85.50\pm2.10$ \\
         $\checkmark$ & - & $\checkmark$ & $89.79\pm2.63$ \\
         $\checkmark$ & $\checkmark$ & $\checkmark$ & $91.54\pm 1.48$ \\
         \bottomrule
    \end{tabular}
    \label{tab:ablation}
\end{table}

\begin{figure}
    \centering
    \includegraphics[width=0.4\textwidth]{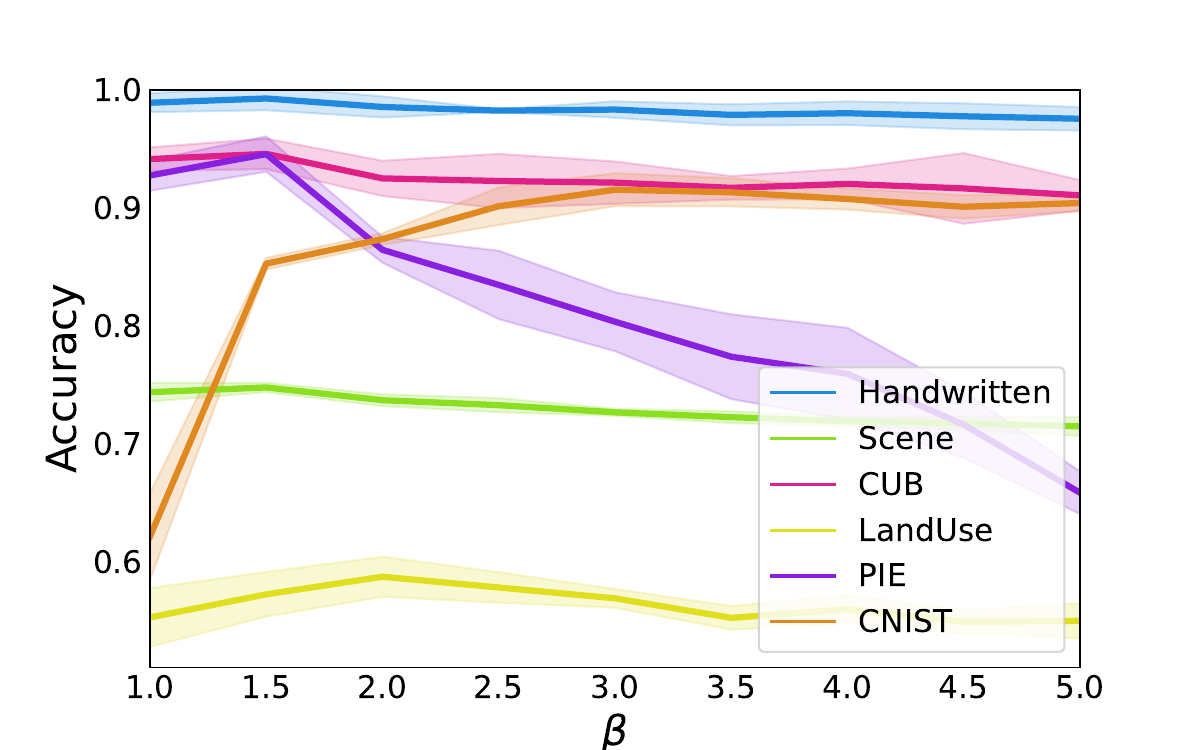}
    \caption{The accuracy with different hyper-parameter $\beta$.}
    \label{fig:parameter}
\end{figure}

\subsubsection{Analysis}

\paragraph{Ablation Study.}To validate the effectiveness of the evidence decoupling strategy, separability principle, and error evidence reduction principle, we construct a detailed ablation study that performs different combinations of these modules to achieve degradation methods. Specifically, to verify the effectiveness of the evidence decoupling module, we compared the approach with the three modules removed to using only that module. The effectiveness of the other two modules was validated based on the decoupling module respectively. We conduct the degradation methods above on the dataset Colored-MNIST. The results are shown in Table \ref{tab:ablation}. From the result, we can observe the outstanding effectiveness of our dynamic decoupling strategy. Therefore, to further validate the effectiveness of the evidence decoupling module, we propose 6 variants of the CCML which Only use Consistent Evidence, Only Complementary Evidence, simple Accumulation of Evidence, simple Average of Evidence, Average of Evidence with Separation, and Accumulate of Evidence with Separation, respectively. We conduct CCML and variants on the CUB datasets and obtain experimental results as shown in Table \ref{tab:CC}. Compared to other methods, CCML achieves higher accuracy because it decouples and processes consistent and complementary evidence separately, giving higher confidence and increasing the separation degree for consistent evidence while averaging complementary evidence. This allows CCML to adjust the corresponding uncertainties based on the consistency between different views, instead of considering only one type of evidence or applying the same separation processing strategy to both consistent and complementary evidence. From the ablation study, we verified the effectiveness of each module of CCML.

\paragraph{Parameter Analysis}The separation increase module can improve the model's performance in solving classification tasks. We verify the influence of the separation increase module on the model by changing the value of the hyperparameter $\beta$. Specifically, we gradually increase the hyperparameter $\beta $ from 1 to 5 and observe CCML's performance on all datasets as shown in Figure \ref{fig:parameter}. The results show that the accuracy of the model increases first and then decreases with the change of $\beta$. In particular, the effectiveness of $\beta$ in the semantic vagueness phenomenon is demonstrated on the dataset Colored-MNIST by the significantly increased accuracy. We can obtain the point that when the $\beta$ is too large, it will have a negative impact on the model, and the appropriate $\beta$ value can improve the performance of the model to a certain extent.

\begin{figure}
    \centering
    \subfloat[$\eta = 0.1$]
      {
          \label{fig:uncertainty1}\includegraphics[width=0.2\textwidth]{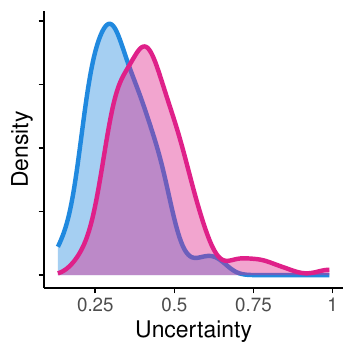}
      }
    \subfloat[$\eta = 1$]
      {
          \label{fig:uncertainty2}\includegraphics[width=0.2\textwidth]{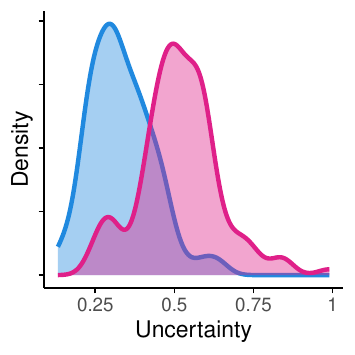}
      }\\
    \subfloat[$\eta = 2$]
      {
          \label{fig:uncertainty3}\includegraphics[width=0.2\textwidth]{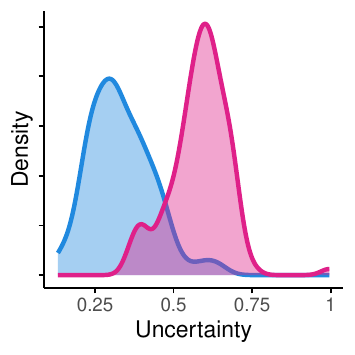}
      }
    \subfloat[$\eta = 5$]
      {
          \label{fig:uncertainty4}\includegraphics[width=0.2\textwidth]{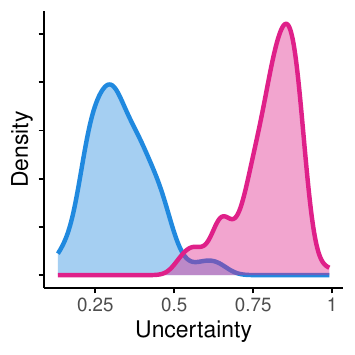}
      }
    \caption{\label{fig:uncertainty}Uncertainty comparison on noise datasets.}
\end{figure}

\paragraph{Uncertainty Estimation.}To further evaluate the estimated uncertainty, we used the original dataset CUB and constructed out-of-distribution instances. We consider the original test instances as in-distribution data and add Gaussian noise $N(0,\boldsymbol{I})$ with intensity $\eta$ to the test instances in one view, constructing out-of-distribution test instances. Specifically, given the noise vector $\boldsymbol{\epsilon}$ sampled from the Gaussian distribution, the out-of-distribution test instances $\boldsymbol{\widetilde{x}}_i = \boldsymbol{x}_i + \eta\boldsymbol{\epsilon}_i$. The uncertainty associated with out-of-distribution data is expected to be higher compared to that of in-distribution data. The noise intensity increases in the sequence $(\eta = 0.1, 1, 2, 5)$. We perform CCML on the data with added Gaussian noise and visualize the uncertainty, as shown in Figure \ref{fig:uncertainty}, where the blue curves represent in-distribution instances and the red curves represent out-of-distribution instances with noise intensity $\eta$. The results show that as the intensity of noise increases, the overall distribution of the uncertainty also increases, demonstrating that the data have higher uncertainty with greater noise. This also demonstrates the ability of our method to estimate uncertainty.

\section{Conclusion}
In this paper, we propose a CCML method to solve the semantic vagueness problem in trusted multi-view learning. CCML tries to dynamically decouple the consistent and complementary evidence from the view-specific evidence. It further processes consistent and complementary evidence according to different principles to achieve classification results and reliability. The experimental results on the synthetic toy dataset and six real-world datasets verified the effectiveness of the proposed decouple strategy and the performance superiority of CCML.

\begin{acks}
This research was supported by the National Natural Science Foundation of China (Grant Nos. 62133012, 61936006, 62103314, 62203354, 62303366) and the Key Research and Development Program of Shaanxi (Program No.2024GX-YBXM-122).
\end{acks}

\bibliographystyle{ACM-Reference-Format}
\balance
\bibliography{acmart}


\begin{thebibliography}{44}

%%% the Web address.

\ifx \showCODEN    \undefined \def \showCODEN     #1{\unskip}     \fi
\ifx \showDOI      \undefined \def \showDOI       #1{#1}\fi
\ifx \showISBNx    \undefined \def \showISBNx     #1{\unskip}     \fi
\ifx \showISBNxiii \undefined \def \showISBNxiii  #1{\unskip}     \fi
\ifx \showISSN     \undefined \def \showISSN      #1{\unskip}     \fi
\ifx \showLCCN     \undefined \def \showLCCN      #1{\unskip}     \fi
\ifx \shownote     \undefined \def \shownote      #1{#1}          \fi
\ifx \showarticletitle \undefined \def \showarticletitle #1{#1}   \fi
\ifx \showURL      \undefined \def \showURL       {\relax}        \fi
% invisible to TeX
\providecommand\bibfield[2]{#2}
\providecommand\bibinfo[2]{#2}
\providecommand\natexlab[1]{#1}
\providecommand\showeprint[2][]{arXiv:#2}

\bibitem[Andrew et~al\mbox{.}(2013)]%
        {andrew2013deep}
\bibfield{author}{\bibinfo{person}{Galen Andrew}, \bibinfo{person}{Raman Arora}, \bibinfo{person}{Jeff Bilmes}, {and} \bibinfo{person}{Karen Livescu}.} \bibinfo{year}{2013}\natexlab{}.
\newblock \showarticletitle{Deep canonical correlation analysis}. In \bibinfo{booktitle}{\emph{International conference on machine learning}}. PMLR, \bibinfo{pages}{1247--1255}.
\newblock


\bibitem[Chao and Sun(2016)]%
        {chao2016consensus}
\bibfield{author}{\bibinfo{person}{Guoqing Chao} {and} \bibinfo{person}{Shiliang Sun}.} \bibinfo{year}{2016}\natexlab{}.
\newblock \showarticletitle{Consensus and complementarity based maximum entropy discrimination for multi-view classification}.
\newblock \bibinfo{journal}{\emph{Information Sciences}}  \bibinfo{volume}{367} (\bibinfo{year}{2016}), \bibinfo{pages}{296--310}.
\newblock


\bibitem[Chen et~al\mbox{.}(2024)]%
        {10440511}
\bibfield{author}{\bibinfo{person}{Changrui Chen}, \bibinfo{person}{Jungong Han}, {and} \bibinfo{person}{Kurt Debattista}.} \bibinfo{year}{2024}\natexlab{}.
\newblock \showarticletitle{Virtual Category Learning: A Semi-Supervised Learning Method for Dense Prediction With Extremely Limited Labels}.
\newblock \bibinfo{journal}{\emph{IEEE Transactions on Pattern Analysis and Machine Intelligence}} \bibinfo{volume}{46}, \bibinfo{number}{8} (\bibinfo{year}{2024}), \bibinfo{pages}{5595--5611}.
\newblock
\urldef\tempurl%
\url{https://doi.org/10.1109/TPAMI.2024.3367416}
\showDOI{\tempurl}


\bibitem[Deng et~al\mbox{.}(2023)]%
        {deng2023uncertainty}
\bibfield{author}{\bibinfo{person}{Danruo Deng}, \bibinfo{person}{Guangyong Chen}, \bibinfo{person}{Yang Yu}, \bibinfo{person}{Furui Liu}, {and} \bibinfo{person}{Pheng-Ann Heng}.} \bibinfo{year}{2023}\natexlab{}.
\newblock \showarticletitle{Uncertainty estimation by fisher information-based evidential deep learning}. In \bibinfo{booktitle}{\emph{International Conference on Machine Learning}}. PMLR, \bibinfo{pages}{7596--7616}.
\newblock


\bibitem[Dong et~al\mbox{.}(2023)]%
        {dong2023cross}
\bibfield{author}{\bibinfo{person}{Zhibin Dong}, \bibinfo{person}{Siwei Wang}, \bibinfo{person}{Jiaqi Jin}, \bibinfo{person}{Xinwang Liu}, {and} \bibinfo{person}{En Zhu}.} \bibinfo{year}{2023}\natexlab{}.
\newblock \showarticletitle{Cross-view Topology Based Consistent and Complementary Information for Deep Multi-view Clustering}. In \bibinfo{booktitle}{\emph{Proceedings of the IEEE/CVF International Conference on Computer Vision}}. \bibinfo{pages}{19440--19451}.
\newblock


\bibitem[Duan et~al\mbox{.}(2018)]%
        {duan2018unified}
\bibfield{author}{\bibinfo{person}{Jiali Duan}, \bibinfo{person}{Jun Wan}, \bibinfo{person}{Shuai Zhou}, \bibinfo{person}{Xiaoyuan Guo}, {and} \bibinfo{person}{Stan~Z Li}.} \bibinfo{year}{2018}\natexlab{}.
\newblock \showarticletitle{A unified framework for multi-modal isolated gesture recognition}.
\newblock \bibinfo{journal}{\emph{ACM Transactions on Multimedia Computing, Communications, and Applications (TOMM)}} \bibinfo{volume}{14}, \bibinfo{number}{1s} (\bibinfo{year}{2018}), \bibinfo{pages}{1--16}.
\newblock


\bibitem[Fang et~al\mbox{.}(2024)]%
        {fang2024representation}
\bibfield{author}{\bibinfo{person}{Zihan Fang}, \bibinfo{person}{Shide Du}, \bibinfo{person}{Zhiling Cai}, \bibinfo{person}{Shiyang Lan}, \bibinfo{person}{Chunming Wu}, \bibinfo{person}{Yanchao Tan}, {and} \bibinfo{person}{Shiping Wang}.} \bibinfo{year}{2024}\natexlab{}.
\newblock \showarticletitle{Representation Learning Meets Optimization-Derived Networks: From Single-View to Multi-View}.
\newblock \bibinfo{journal}{\emph{IEEE Transactions on Multimedia}} (\bibinfo{year}{2024}).
\newblock


\bibitem[Gal and Ghahramani(2016)]%
        {gal2016dropout}
\bibfield{author}{\bibinfo{person}{Yarin Gal} {and} \bibinfo{person}{Zoubin Ghahramani}.} \bibinfo{year}{2016}\natexlab{}.
\newblock \showarticletitle{Dropout as a bayesian approximation: Representing model uncertainty in deep learning}. In \bibinfo{booktitle}{\emph{international conference on machine learning}}. PMLR, \bibinfo{pages}{1050--1059}.
\newblock


\bibitem[Gan et~al\mbox{.}(2021)]%
        {gan2021brain}
\bibfield{author}{\bibinfo{person}{Jiangzhang Gan}, \bibinfo{person}{Ziwen Peng}, \bibinfo{person}{Xiaofeng Zhu}, \bibinfo{person}{Rongyao Hu}, \bibinfo{person}{Junbo Ma}, {and} \bibinfo{person}{Guorong Wu}.} \bibinfo{year}{2021}\natexlab{}.
\newblock \showarticletitle{Brain functional connectivity analysis based on multi-graph fusion}.
\newblock \bibinfo{journal}{\emph{Medical image analysis}}  \bibinfo{volume}{71} (\bibinfo{year}{2021}), \bibinfo{pages}{102057}.
\newblock


\bibitem[Han et~al\mbox{.}(2021)]%
        {han2021tmc}
\bibfield{author}{\bibinfo{person}{Zongbo Han}, \bibinfo{person}{Changqing Zhang}, \bibinfo{person}{Huazhu Fu}, {and} \bibinfo{person}{Joey~Tianyi Zhou}.} \bibinfo{year}{2021}\natexlab{}.
\newblock \showarticletitle{Trusted multi-view classification}. In \bibinfo{booktitle}{\emph{International Conference on Learning Representations}}.
\newblock


\bibitem[Han et~al\mbox{.}(2022)]%
        {han2022trusted}
\bibfield{author}{\bibinfo{person}{Zongbo Han}, \bibinfo{person}{Changqing Zhang}, \bibinfo{person}{Huazhu Fu}, {and} \bibinfo{person}{Joey~Tianyi Zhou}.} \bibinfo{year}{2022}\natexlab{}.
\newblock \showarticletitle{Trusted multi-view classification with dynamic evidential fusion}.
\newblock \bibinfo{journal}{\emph{IEEE transactions on pattern analysis and machine intelligence}} \bibinfo{volume}{45}, \bibinfo{number}{2} (\bibinfo{year}{2022}), \bibinfo{pages}{2551--2566}.
\newblock


\bibitem[Hendrycks and Gimpel(2016)]%
        {hendrycks2016baseline}
\bibfield{author}{\bibinfo{person}{Dan Hendrycks} {and} \bibinfo{person}{Kevin Gimpel}.} \bibinfo{year}{2016}\natexlab{}.
\newblock \showarticletitle{A baseline for detecting misclassified and out-of-distribution examples in neural networks}.
\newblock \bibinfo{journal}{\emph{arXiv preprint arXiv:1610.02136}} (\bibinfo{year}{2016}).
\newblock


\bibitem[Hu et~al\mbox{.}(2024)]%
        {hu2024deep}
\bibfield{author}{\bibinfo{person}{Shizhe Hu}, \bibinfo{person}{Chengkun Zhang}, \bibinfo{person}{Guoliang Zou}, \bibinfo{person}{Zhengzheng Lou}, {and} \bibinfo{person}{Yangdong Ye}.} \bibinfo{year}{2024}\natexlab{}.
\newblock \showarticletitle{Deep Multiview Clustering by Pseudo-Label Guided Contrastive Learning and Dual Correlation Learning}.
\newblock \bibinfo{journal}{\emph{IEEE Transactions on Neural Networks and Learning Systems}} (\bibinfo{year}{2024}).
\newblock


\bibitem[Huang et~al\mbox{.}(2020)]%
        {huang2020auto}
\bibfield{author}{\bibinfo{person}{Shudong Huang}, \bibinfo{person}{Zhao Kang}, {and} \bibinfo{person}{Zenglin Xu}.} \bibinfo{year}{2020}\natexlab{}.
\newblock \showarticletitle{Auto-weighted multi-view clustering via deep matrix decomposition}.
\newblock \bibinfo{journal}{\emph{Pattern Recognition}}  \bibinfo{volume}{97} (\bibinfo{year}{2020}), \bibinfo{pages}{107015}.
\newblock


\bibitem[Jillani et~al\mbox{.}(2020)]%
        {jillani2020multi}
\bibfield{author}{\bibinfo{person}{Rashad Jillani}, \bibinfo{person}{Syed~Fawad Hussain}, {and} \bibinfo{person}{Hari Kalva}.} \bibinfo{year}{2020}\natexlab{}.
\newblock \showarticletitle{Multi-view clustering for fast intra mode decision in HEVC}. In \bibinfo{booktitle}{\emph{2020 IEEE International Conference on Consumer Electronics (ICCE)}}. IEEE, \bibinfo{pages}{1--4}.
\newblock


\bibitem[J{\o}sang(2016)]%
        {josang2016subjective}
\bibfield{author}{\bibinfo{person}{Audun J{\o}sang}.} \bibinfo{year}{2016}\natexlab{}.
\newblock \bibinfo{booktitle}{\emph{Subjective logic}}. Vol.~\bibinfo{volume}{3}.
\newblock \bibinfo{publisher}{Springer}.
\newblock


\bibitem[Kingma and Ba(2014)]%
        {kingma2014adam}
\bibfield{author}{\bibinfo{person}{Diederik~P Kingma} {and} \bibinfo{person}{Jimmy Ba}.} \bibinfo{year}{2014}\natexlab{}.
\newblock \showarticletitle{Adam: A method for stochastic optimization}.
\newblock \bibinfo{journal}{\emph{arXiv preprint arXiv:1412.6980}} (\bibinfo{year}{2014}).
\newblock


\bibitem[Liang et~al\mbox{.}(2021)]%
        {liang2021af}
\bibfield{author}{\bibinfo{person}{Xinyan Liang}, \bibinfo{person}{Yuhua Qian}, \bibinfo{person}{Qian Guo}, \bibinfo{person}{Honghong Cheng}, {and} \bibinfo{person}{Jiye Liang}.} \bibinfo{year}{2021}\natexlab{}.
\newblock \showarticletitle{AF: An association-based fusion method for multi-modal classification}.
\newblock \bibinfo{journal}{\emph{IEEE Transactions on Pattern Analysis and Machine Intelligence}} \bibinfo{volume}{44}, \bibinfo{number}{12} (\bibinfo{year}{2021}), \bibinfo{pages}{9236--9254}.
\newblock


\bibitem[Lin et~al\mbox{.}(2022)]%
        {lin2022dual}
\bibfield{author}{\bibinfo{person}{Yijie Lin}, \bibinfo{person}{Yuanbiao Gou}, \bibinfo{person}{Xiaotian Liu}, \bibinfo{person}{Jinfeng Bai}, \bibinfo{person}{Jiancheng Lv}, {and} \bibinfo{person}{Xi Peng}.} \bibinfo{year}{2022}\natexlab{}.
\newblock \showarticletitle{Dual contrastive prediction for incomplete multi-view representation learning}.
\newblock \bibinfo{journal}{\emph{IEEE Transactions on Pattern Analysis and Machine Intelligence}} \bibinfo{volume}{45}, \bibinfo{number}{4} (\bibinfo{year}{2022}), \bibinfo{pages}{4447--4461}.
\newblock


\bibitem[Liu et~al\mbox{.}(2024)]%
        {liu2024attention}
\bibfield{author}{\bibinfo{person}{Chengliang Liu}, \bibinfo{person}{Jinlong Jia}, \bibinfo{person}{Jie Wen}, \bibinfo{person}{Yabo Liu}, \bibinfo{person}{Xiaoling Luo}, \bibinfo{person}{Chao Huang}, {and} \bibinfo{person}{Yong Xu}.} \bibinfo{year}{2024}\natexlab{}.
\newblock \showarticletitle{Attention-Induced Embedding Imputation for Incomplete Multi-View Partial Multi-Label Classification}. In \bibinfo{booktitle}{\emph{Proceedings of the AAAI Conference on Artificial Intelligence}}, Vol.~\bibinfo{volume}{38}. \bibinfo{pages}{13864--13872}.
\newblock


\bibitem[Liu et~al\mbox{.}(2022a)]%
        {liu2022disentangled}
\bibfield{author}{\bibinfo{person}{Fan Liu}, \bibinfo{person}{Huilin Chen}, \bibinfo{person}{Zhiyong Cheng}, \bibinfo{person}{Anan Liu}, \bibinfo{person}{Liqiang Nie}, {and} \bibinfo{person}{Mohan Kankanhalli}.} \bibinfo{year}{2022}\natexlab{a}.
\newblock \showarticletitle{Disentangled multimodal representation learning for recommendation}.
\newblock \bibinfo{journal}{\emph{IEEE Transactions on Multimedia}} (\bibinfo{year}{2022}).
\newblock


\bibitem[Liu et~al\mbox{.}(2022b)]%
        {liu2022trusted}
\bibfield{author}{\bibinfo{person}{Wei Liu}, \bibinfo{person}{Xiaodong Yue}, \bibinfo{person}{Yufei Chen}, {and} \bibinfo{person}{Thierry Denoeux}.} \bibinfo{year}{2022}\natexlab{b}.
\newblock \showarticletitle{Trusted multi-view deep learning with opinion aggregation}. In \bibinfo{booktitle}{\emph{Proceedings of the AAAI Conference on Artificial Intelligence}}, Vol.~\bibinfo{volume}{36}. \bibinfo{pages}{7585--7593}.
\newblock


\bibitem[Min et~al\mbox{.}(2023)]%
        {min2023recent}
\bibfield{author}{\bibinfo{person}{Bonan Min}, \bibinfo{person}{Hayley Ross}, \bibinfo{person}{Elior Sulem}, \bibinfo{person}{Amir Pouran~Ben Veyseh}, \bibinfo{person}{Thien~Huu Nguyen}, \bibinfo{person}{Oscar Sainz}, \bibinfo{person}{Eneko Agirre}, \bibinfo{person}{Ilana Heintz}, {and} \bibinfo{person}{Dan Roth}.} \bibinfo{year}{2023}\natexlab{}.
\newblock \showarticletitle{Recent advances in natural language processing via large pre-trained language models: A survey}.
\newblock \bibinfo{journal}{\emph{Comput. Surveys}} \bibinfo{volume}{56}, \bibinfo{number}{2} (\bibinfo{year}{2023}), \bibinfo{pages}{1--40}.
\newblock


\bibitem[Morvant et~al\mbox{.}(2014)]%
        {morvant2014majority}
\bibfield{author}{\bibinfo{person}{Emilie Morvant}, \bibinfo{person}{Amaury Habrard}, {and} \bibinfo{person}{St{\'e}phane Ayache}.} \bibinfo{year}{2014}\natexlab{}.
\newblock \showarticletitle{Majority vote of diverse classifiers for late fusion}. In \bibinfo{booktitle}{\emph{Structural, Syntactic, and Statistical Pattern Recognition: Joint IAPR International Workshop, S+ SSPR 2014, Joensuu, Finland, August 20-22, 2014. Proceedings}}. Springer, \bibinfo{pages}{153--162}.
\newblock


\bibitem[Neal(2012)]%
        {neal2012bayesian}
\bibfield{author}{\bibinfo{person}{Radford~M Neal}.} \bibinfo{year}{2012}\natexlab{}.
\newblock \bibinfo{booktitle}{\emph{Bayesian learning for neural networks}}. Vol.~\bibinfo{volume}{118}.
\newblock \bibinfo{publisher}{Springer Science \& Business Media}.
\newblock


\bibitem[Qin et~al\mbox{.}(2022)]%
        {qin2022deep}
\bibfield{author}{\bibinfo{person}{Yang Qin}, \bibinfo{person}{Dezhong Peng}, \bibinfo{person}{Xi Peng}, \bibinfo{person}{Xu Wang}, {and} \bibinfo{person}{Peng Hu}.} \bibinfo{year}{2022}\natexlab{}.
\newblock \showarticletitle{Deep evidential learning with noisy correspondence for cross-modal retrieval}. In \bibinfo{booktitle}{\emph{Proceedings of the 30th ACM International Conference on Multimedia}}. \bibinfo{pages}{4948--4956}.
\newblock


\bibitem[Roitberg et~al\mbox{.}(2019)]%
        {roitberg2019analysis}
\bibfield{author}{\bibinfo{person}{Alina Roitberg}, \bibinfo{person}{Tim Pollert}, \bibinfo{person}{Monica Haurilet}, \bibinfo{person}{Manuel Martin}, {and} \bibinfo{person}{Rainer Stiefelhagen}.} \bibinfo{year}{2019}\natexlab{}.
\newblock \showarticletitle{Analysis of deep fusion strategies for multi-modal gesture recognition}. In \bibinfo{booktitle}{\emph{Proceedings of the IEEE/CVF Conference on Computer Vision and Pattern Recognition Workshops}}. \bibinfo{pages}{0--0}.
\newblock


\bibitem[Sensoy et~al\mbox{.}(2018)]%
        {sensoy2018evidential}
\bibfield{author}{\bibinfo{person}{Murat Sensoy}, \bibinfo{person}{Lance Kaplan}, {and} \bibinfo{person}{Melih Kandemir}.} \bibinfo{year}{2018}\natexlab{}.
\newblock \showarticletitle{Evidential deep learning to quantify classification uncertainty}.
\newblock \bibinfo{journal}{\emph{Advances in neural information processing systems}}  \bibinfo{volume}{31} (\bibinfo{year}{2018}).
\newblock


\bibitem[Shutova et~al\mbox{.}(2016)]%
        {shutova2016black}
\bibfield{author}{\bibinfo{person}{Ekaterina Shutova}, \bibinfo{person}{Douwe Kiela}, {and} \bibinfo{person}{Jean Maillard}.} \bibinfo{year}{2016}\natexlab{}.
\newblock \showarticletitle{Black holes and white rabbits: Metaphor identification with visual features}. In \bibinfo{booktitle}{\emph{Proceedings of the 2016 conference of the North American chapter of the association for computational linguistics: Human language technologies}}. \bibinfo{pages}{160--170}.
\newblock


\bibitem[Sun et~al\mbox{.}(2021)]%
        {sun2021scalable}
\bibfield{author}{\bibinfo{person}{Mengjing Sun}, \bibinfo{person}{Pei Zhang}, \bibinfo{person}{Siwei Wang}, \bibinfo{person}{Sihang Zhou}, \bibinfo{person}{Wenxuan Tu}, \bibinfo{person}{Xinwang Liu}, \bibinfo{person}{En Zhu}, {and} \bibinfo{person}{Changjian Wang}.} \bibinfo{year}{2021}\natexlab{}.
\newblock \showarticletitle{Scalable multi-view subspace clustering with unified anchors}. In \bibinfo{booktitle}{\emph{Proceedings of the 29th ACM International Conference on Multimedia}}. \bibinfo{pages}{3528--3536}.
\newblock


\bibitem[Wah et~al\mbox{.}(2011)]%
        {wah2011caltech}
\bibfield{author}{\bibinfo{person}{Catherine Wah}, \bibinfo{person}{Steve Branson}, \bibinfo{person}{Peter Welinder}, \bibinfo{person}{Pietro Perona}, {and} \bibinfo{person}{Serge Belongie}.} \bibinfo{year}{2011}\natexlab{}.
\newblock \showarticletitle{The caltech-ucsd birds-200-2011 dataset}.
\newblock  (\bibinfo{year}{2011}).
\newblock


\bibitem[Wang et~al\mbox{.}(2015)]%
        {wang2015deep}
\bibfield{author}{\bibinfo{person}{Weiran Wang}, \bibinfo{person}{Raman Arora}, \bibinfo{person}{Karen Livescu}, {and} \bibinfo{person}{Jeff Bilmes}.} \bibinfo{year}{2015}\natexlab{}.
\newblock \showarticletitle{On deep multi-view representation learning}. In \bibinfo{booktitle}{\emph{International conference on machine learning}}. PMLR, \bibinfo{pages}{1083--1092}.
\newblock


\bibitem[Wei et~al\mbox{.}(2024)]%
        {wei2024video}
\bibfield{author}{\bibinfo{person}{Kaiwen Wei}, \bibinfo{person}{Runyan Du}, \bibinfo{person}{Li Jin}, \bibinfo{person}{Jian Liu}, \bibinfo{person}{Jianhua Yin}, \bibinfo{person}{Linhao Zhang}, \bibinfo{person}{Jintao Liu}, \bibinfo{person}{Nayu Liu}, \bibinfo{person}{Jingyuan Zhang}, {and} \bibinfo{person}{Zhi Guo}.} \bibinfo{year}{2024}\natexlab{}.
\newblock \showarticletitle{Video Event Extraction with Multi-View Interaction Knowledge Distillation}. In \bibinfo{booktitle}{\emph{Proceedings of the AAAI Conference on Artificial Intelligence}}, Vol.~\bibinfo{volume}{38}. \bibinfo{pages}{19224--19233}.
\newblock


\bibitem[Xie et~al\mbox{.}(2023)]%
        {xie2023exploring}
\bibfield{author}{\bibinfo{person}{Mengyao Xie}, \bibinfo{person}{Zongbo Han}, \bibinfo{person}{Changqing Zhang}, \bibinfo{person}{Yichen Bai}, {and} \bibinfo{person}{Qinghua Hu}.} \bibinfo{year}{2023}\natexlab{}.
\newblock \showarticletitle{Exploring and exploiting uncertainty for incomplete multi-view classification}. In \bibinfo{booktitle}{\emph{Proceedings of the IEEE/CVF Conference on Computer Vision and Pattern Recognition}}. \bibinfo{pages}{19873--19882}.
\newblock


\bibitem[Xu et~al\mbox{.}(2018)]%
        {xu2018deep}
\bibfield{author}{\bibinfo{person}{Cai Xu}, \bibinfo{person}{Ziyu Guan}, \bibinfo{person}{Wei Zhao}, \bibinfo{person}{Yunfei Niu}, \bibinfo{person}{Quan Wang}, {and} \bibinfo{person}{Zhiheng Wang}.} \bibinfo{year}{2018}\natexlab{}.
\newblock \showarticletitle{Deep multi-view concept learning.}. In \bibinfo{booktitle}{\emph{IJCAI}}. Stockholm, \bibinfo{pages}{2898--2904}.
\newblock


\bibitem[Xu et~al\mbox{.}(2024)]%
        {xu2024reliable}
\bibfield{author}{\bibinfo{person}{Cai Xu}, \bibinfo{person}{Jiajun Si}, \bibinfo{person}{Ziyu Guan}, \bibinfo{person}{Wei Zhao}, \bibinfo{person}{Yue Wu}, {and} \bibinfo{person}{Xiyue Gao}.} \bibinfo{year}{2024}\natexlab{}.
\newblock \showarticletitle{Reliable Conflictive Multi-View Learning}.
\newblock \bibinfo{journal}{\emph{Proceedings of the AAAI Conference on Artificial Intelligence}} \bibinfo{volume}{38}, \bibinfo{number}{14} (\bibinfo{date}{Mar.} \bibinfo{year}{2024}), \bibinfo{pages}{16129--16137}.
\newblock
\urldef\tempurl%
\url{https://doi.org/10.1609/aaai.v38i14.29546}
\showDOI{\tempurl}


\bibitem[Xu et~al\mbox{.}(2023)]%
        {9906001}
\bibfield{author}{\bibinfo{person}{Cai Xu}, \bibinfo{person}{Wei Zhao}, \bibinfo{person}{Jinglong Zhao}, \bibinfo{person}{Ziyu Guan}, \bibinfo{person}{Xiangyu Song}, {and} \bibinfo{person}{Jianxin Li}.} \bibinfo{year}{2023}\natexlab{}.
\newblock \showarticletitle{Uncertainty-Aware Multiview Deep Learning for Internet of Things Applications}.
\newblock \bibinfo{journal}{\emph{IEEE Transactions on Industrial Informatics}} \bibinfo{volume}{19}, \bibinfo{number}{2} (\bibinfo{year}{2023}), \bibinfo{pages}{1456--1466}.
\newblock
\urldef\tempurl%
\url{https://doi.org/10.1109/TII.2022.3206343}
\showDOI{\tempurl}


\bibitem[Xu et~al\mbox{.}(2021)]%
        {xu2021multi}
\bibfield{author}{\bibinfo{person}{Jie Xu}, \bibinfo{person}{Yazhou Ren}, \bibinfo{person}{Huayi Tang}, \bibinfo{person}{Xiaorong Pu}, \bibinfo{person}{Xiaofeng Zhu}, \bibinfo{person}{Ming Zeng}, {and} \bibinfo{person}{Lifang He}.} \bibinfo{year}{2021}\natexlab{}.
\newblock \showarticletitle{Multi-VAE: Learning disentangled view-common and view-peculiar visual representations for multi-view clustering}. In \bibinfo{booktitle}{\emph{Proceedings of the IEEE/CVF international conference on computer vision}}. \bibinfo{pages}{9234--9243}.
\newblock


\bibitem[Yang and Newsam(2010)]%
        {yang2010bag}
\bibfield{author}{\bibinfo{person}{Yi Yang} {and} \bibinfo{person}{Shawn Newsam}.} \bibinfo{year}{2010}\natexlab{}.
\newblock \showarticletitle{Bag-of-visual-words and spatial extensions for land-use classification}. In \bibinfo{booktitle}{\emph{Proceedings of the 18th SIGSPATIAL international conference on advances in geographic information systems}}. \bibinfo{pages}{270--279}.
\newblock


\bibitem[Yin and Sun(2021)]%
        {yin2021incomplete}
\bibfield{author}{\bibinfo{person}{Jun Yin} {and} \bibinfo{person}{Shiliang Sun}.} \bibinfo{year}{2021}\natexlab{}.
\newblock \showarticletitle{Incomplete multi-view clustering with reconstructed views}.
\newblock \bibinfo{journal}{\emph{IEEE Transactions on Knowledge and Data Engineering}} \bibinfo{volume}{35}, \bibinfo{number}{3} (\bibinfo{year}{2021}), \bibinfo{pages}{2671--2682}.
\newblock


\bibitem[Zhang et~al\mbox{.}(2019)]%
        {NEURIPS2019_11b9842e}
\bibfield{author}{\bibinfo{person}{Changqing Zhang}, \bibinfo{person}{Zongbo Han}, \bibinfo{person}{yajie cui}, \bibinfo{person}{Huazhu Fu}, \bibinfo{person}{Joey~Tianyi Zhou}, {and} \bibinfo{person}{Qinghua Hu}.} \bibinfo{year}{2019}\natexlab{}.
\newblock \showarticletitle{CPM-Nets: Cross Partial Multi-View Networks}. In \bibinfo{booktitle}{\emph{Advances in Neural Information Processing Systems}}, \bibfield{editor}{\bibinfo{person}{H.~Wallach}, \bibinfo{person}{H.~Larochelle}, \bibinfo{person}{A.~Beygelzimer}, \bibinfo{person}{F.~d\textquotesingle Alch\'{e}-Buc}, \bibinfo{person}{E.~Fox}, {and} \bibinfo{person}{R.~Garnett}} (Eds.), Vol.~\bibinfo{volume}{32}. \bibinfo{publisher}{Curran Associates, Inc.}
\newblock
\urldef\tempurl%
\url{https://proceedings.neurips.cc/paper/2019/file/11b9842e0a271ff252c1903e7132cd68-Paper.pdf}
\showURL{%
\tempurl}


\bibitem[Zhang et~al\mbox{.}(2024)]%
        {zhang2024multimodalfusionlowqualitydata}
\bibfield{author}{\bibinfo{person}{Qingyang Zhang}, \bibinfo{person}{Yake Wei}, \bibinfo{person}{Zongbo Han}, \bibinfo{person}{Huazhu Fu}, \bibinfo{person}{Xi Peng}, \bibinfo{person}{Cheng Deng}, \bibinfo{person}{Qinghua Hu}, \bibinfo{person}{Cai Xu}, \bibinfo{person}{Jie Wen}, \bibinfo{person}{Di Hu}, {and} \bibinfo{person}{Changqing Zhang}.} \bibinfo{year}{2024}\natexlab{}.
\newblock \bibinfo{title}{Multimodal Fusion on Low-quality Data: A Comprehensive Survey}.
\newblock
\newblock
\showeprint[arxiv]{2404.18947}~[cs.LG]
\urldef\tempurl%
\url{https://arxiv.org/abs/2404.18947}
\showURL{%
\tempurl}


\bibitem[Zhen et~al\mbox{.}(2019)]%
        {zhen2019deep}
\bibfield{author}{\bibinfo{person}{Liangli Zhen}, \bibinfo{person}{Peng Hu}, \bibinfo{person}{Xu Wang}, {and} \bibinfo{person}{Dezhong Peng}.} \bibinfo{year}{2019}\natexlab{}.
\newblock \showarticletitle{Deep supervised cross-modal retrieval}. In \bibinfo{booktitle}{\emph{Proceedings of the IEEE/CVF Conference on Computer Vision and Pattern Recognition}}. \bibinfo{pages}{10394--10403}.
\newblock


\bibitem[Zhou et~al\mbox{.}(2023)]%
        {zhou2023calm}
\bibfield{author}{\bibinfo{person}{Hai Zhou}, \bibinfo{person}{Zhe Xue}, \bibinfo{person}{Ying Liu}, \bibinfo{person}{Boang Li}, \bibinfo{person}{Junping Du}, \bibinfo{person}{Meiyu Liang}, {and} \bibinfo{person}{Yuankai Qi}.} \bibinfo{year}{2023}\natexlab{}.
\newblock \showarticletitle{CALM: An Enhanced Encoding and Confidence Evaluating Framework for Trustworthy Multi-view Learning}. In \bibinfo{booktitle}{\emph{Proceedings of the 31st ACM International Conference on Multimedia}}. \bibinfo{pages}{3108--3116}.
\newblock


\end{thebibliography}

% \appendix









\end{document}